\documentclass[3p]{elsarticle}

\usepackage{times}
\usepackage{epsfig}
\usepackage{graphicx}
\usepackage{amsmath}
\usepackage{amssymb}

\usepackage{bm}
\usepackage[dvipsnames]{xcolor}
\usepackage[hypcap=false]{caption, subcaption}
\usepackage{enumitem}
\usepackage{calc}
\usepackage{csquotes}
\usepackage{booktabs}
\usepackage[flushleft]{threeparttable}
\usepackage{mathtools}
\usepackage{multirow}
\usepackage{placeins}
\usepackage{xparse}
\usepackage{xspace}
\usepackage{afterpage}
\usepackage{footnote}
\makesavenoteenv{tabular}
\makesavenoteenv{table}
\usepackage{refcount}
\usepackage{stfloats}
\usepackage{array}
\usepackage{tikz}
\usepackage{wrapfig}
\usepackage{lscape}
\xspaceaddexceptions{$\downarrow$}

\usepackage[pdfauthor=author,pagebackref=true,breaklinks=true,colorlinks,bookmarks=false]{hyperref}
\usepackage[capitalize]{cleveref}  
\crefname{table}{Table}{Tables}

\graphicspath{{img/}}

\iftrue
    \newcommand{\OV}[1]{\textcolor{orange}{OV: #1}}
    \newcommand{\DZ}[1]{\textcolor{blue}{DZ: #1}} 
    \newcommand{\EB}[1]{\textcolor{magenta}{EB: #1}}
    \newcommand{\MK}[1]{\textcolor{LimeGreen}{MK: #1}}
    \newcommand{\AS}[1]{\textcolor{Plum}{AS: #1}}
    \newcommand{\LA}[1]{\textcolor{cyan}{AA: #1}}
    \newcommand{\RV}[1]{}
\else
    \newcommand{\OV}[1]{\textcolor{orange}{}}
    \newcommand{\DZ}[1]{\textcolor{blue}{}} 
    \newcommand{\EB}[1]{\textcolor{magenta}{}}
    \newcommand{\MK}[1]{\textcolor{LimeGreen}{}}
    \newcommand{\AS}[1]{\textcolor{Plum}{}}
    \newcommand{\LA}[1]{\textcolor{cyan}{}}
    \newcommand{\RV}[1]{}
\fi

\DeclarePairedDelimiter\norm{\lVert}{\rVert}

\def\image{I}
\def\dmap{D}
\def\nmap{N}
\def\src{L}
\def\srcup{\src_{\uparrow}}
\def\tgt{H}
\def\tgtdown{\tgt_{\downarrow}}
\def\pseudosrc{\src_{P}}
\def\dmapSrc{\dmap^{\src}}
\def\dmapTgt{\dmap^{\tgt}}

\def\dmapTgtEst{\widehat{\dmap}^{\tgt}}
\def\reconstr{\text{rec}}

\def\dmapTgtdownRec{\widehat{\dmap}^{\tgtdown}_{\reconstr}}

\def\enhanced{\text{enh}}

\def\sr{\text{sr}}
\def\rgb{\text{rgb}}
\def\feedforward{f}

\def\fenh{\feedforward_{\enhanced}}
\def\operation{u}
\def\opsr{\operation_{\sr}}
\def\openh{\operation_{\enhanced}}
\def\opup{\operation_{\text{up}}}

\def\Gen{g}
\def\fimg{\feedforward_{\rgb}}

\def\GenTgtSrc{\Gen_{\tgt 2 \src}}
\def\GenSrcTgt{\Gen_{\src 2 \tgt}}
\def\disc{d}
\def\depth{\text{depth}}
\def\normals{\text{norm}}
\def\discDmapTgt{\disc_{\depth}^{\tgt}}
\def\discNmapTgt{\disc_{\normals}^{\tgt}}
\def\discDmapSrc{\disc_{\depth}^{\src}}
\def\discNmapSrc{\disc_{\normals}^{\src}}
\def\perceptual{\text{surf}}
\def\edge{\text{edge}}
\def\smooth{\text{smooth}}

\def\Loss{\mathcal{L}}
\def\LossDepth{\Loss_{\depth}}

\def\LossPerceptual{\Loss_{\perceptual}}

\def\Reg{\mathcal{R}}
\def\RegSmooth{\Reg_{\smooth}}
\def\RegEdge{\Reg_{\edge}}

\def\lightDir{e}

\NewDocumentCommand\rmse{O{}}{%
    \ifmmode \text{RMSE}_{\text{#1}} 
    \else RMSE\textsubscript{#1}\xspace 
    \fi %
}
\NewDocumentCommand\mse{O{}}{%
    \ifmmode \text{MSE}_{\text{#1}} 
    \else MSE\textsubscript{#1}\xspace 
    \fi %
}
\NewDocumentCommand\mae{O{}}{%
    \ifmmode \text{MAE}_{\text{#1}} 
    \else MAE\textsubscript{#1}\xspace 
    \fi %
}

\xspaceaddexceptions{$\downarrow$}

\makeatletter
\DeclareRobustCommand\onedot{\futurelet\@let@token\@onedot}
\def\@onedot{\ifx\@let@token.\else.\null\fi\xspace}

\def\eg{\emph{e.g}\onedot} 
\def\ie{\emph{i.e}\onedot} 
\def\cf{\emph{c.f}\onedot} 
 
\def\wrt{w.r.t\onedot} 
\def\etal{\emph{et al}\onedot}
\makeatother

\begin{document}
    \title{Unpaired Depth Super-Resolution in the Wild}
\author{\
Aleksandr Safin\textsuperscript{1},
Maxim Kan\textsuperscript{1}\textsuperscript{$\star$},
Nikita Drobyshev\textsuperscript{1}\textsuperscript{$\star$},
Oleg Voynov\textsuperscript{1},
\\\
Alexey Artemov\textsuperscript{1},
Alexander Filippov\textsuperscript{2},
Denis Zorin\textsuperscript{3,1},
Evgeny Burnaev\textsuperscript{1}
\\\
\textsuperscript{1} Skolkovo Institute of Science and Technology,
\textsuperscript{2} Huawei Noah's Ark Lab,
\\\
\textsuperscript{3} New York University
\\\
{\tt\small \{aleksandr.safin, maxim.kan, nikita.drobyshev, oleg.voinov\}@skoltech.ru,\\\ a.artemov@skoltech.ru,   filippov.alexander@huawei.com, dzorin@cs.nyu.edu, e.burnaev@skoltech.ru}
}

\makeatletter{
\renewcommand*{\@makefnmark}{}\footnotetext{\textsuperscript{$\star$}Joint second author contribution.}}

    \begin{abstract}
Depth images captured with commodity sensors commonly suffer from low quality and resolution and require enhancing to be used in many applications.
State-of-the-art data-driven methods for depth super-resolution rely on registered pairs of low- and high-resolution depth images of the same scenes.
Acquisition of such real-world paired data requires specialized setups. 
On the other hand, generating low-resolution depth images from respective high-resolution versions by subsampling, adding noise and other artificial degradation methods, does not fully capture the characteristics of real-world depth data.  
As a consequence, supervised learning methods trained on such artificial paired data may not perform well on real-world low-resolution inputs. 
We propose an approach to depth super-resolution based on learning from \emph{unpaired data}. 
We show that image-based unpaired techniques that have been proposed for depth super-resolution fail to perform effective hole-filling or reconstruct accurate surface normals in the output depth images.
Aiming to improve upon these approaches, we propose an unpaired learning method for depth super-resolution based on a learnable degradation model and including a dedicated enhancement component which integrates surface quality measures to produce more accurate depth images. 
We propose a benchmark for unpaired depth super-resolution and demonstrate that our method outperforms existing unpaired methods and performs on par with paired ones. 
\end{abstract}
\maketitle
\begin{center}
    \centerline{\begin{tikzpicture}
        \node[inner sep=0] (image) {\includegraphics[width=.95\linewidth]{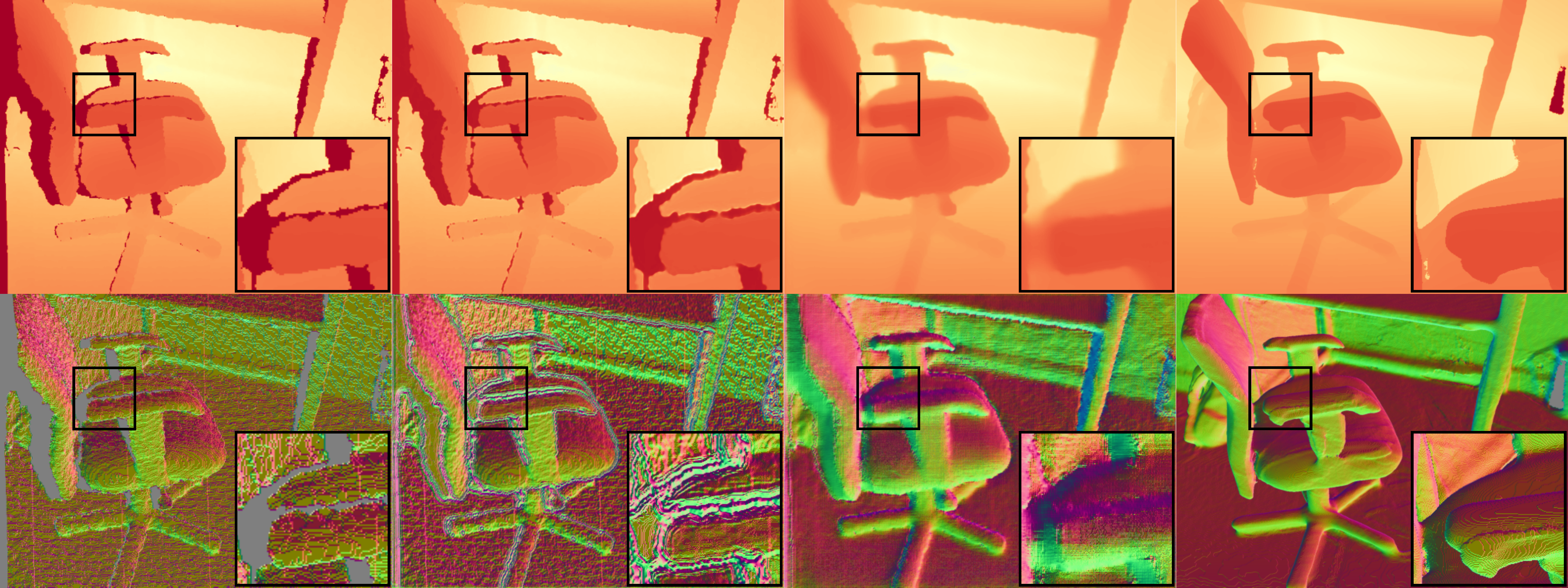}};
        \begin{scope}[shift=(image.north west), x={(image.north east)},y={(image.south west)}]
            \node at (.15, -.04) {Raw Input};
            \node at (.38, -.04) {MS-PFL \cite{MS-PFL:Chuhua2020}};
            \node at (.62, -.04) {Ours};
            \node at (.87, -.04) {Reference};
            \node[rotate=90] at (-.02, .25) {Depth};
            \node[rotate=90] at (-.02, .75) {Surface Normals};
            \tikzstyle{label}=[fill=black, text=white, inner sep=2pt, fill opacity=0.6, text opacity=1]
            \node at (.34, 0.45) [label] {\rmse[d] = 57 mm};
            \node at (.59, 0.45) [label] {\rmse[d] = 54 mm};
            \node at (.33, 0.95) [label] {\mse[v] = 0.31};
            \node at (.58, 0.95) [label] {\mse[v] = 0.24};
        \end{scope}
	\end{tikzpicture}}
    \vspace{-0.5em}
    \captionof{figure}{
    State-of-the-art depth super-resolution methods are designed for clean and complete images, but produce noisy, incomplete results in the wild.
    In contrast, our novel \emph{unpaired} super-resolution method inpaints holes and produces normals closer (\mse[v]~\cite{PDSR:Voynov2019}) to the reference data. 
    \rmse[d] is the depth error averaged over the area of valid (non-hole) pixels.
    }
	\label{fig:sr_fails}
    \vspace{-0.5em}
\end{center}


\section{Introduction}

Depth images are commonly used in a variety of applications, from 3D scene reconstruction to robotic navigation, user interfaces and photo effects. 
Depth sensors are becoming standard for everyday devices such as phones and tablets, immensely expanding availability of this type of data and the range of its applications. 

However, when acquired with  commodity depth cameras, raw depth images come with multiple limitations, most importantly, limited spatial resolution, severe noise levels, and many gaps.
Reliably addressing these flaws has been attracting increasing research interest aimed at enhancing the resolution and quality of depth images.

Following image processing, where the most success has been achieved through deep learning~\cite{DBPN:Haris_2018, SR_review:Yang_2019, SRFBN:Li_2019,TTSR:Yang_2020, SPSR_GG:Ma_2020}, convolutional neural networks (CNNs) were applied to depth super-resolution and enhancement~\cite{ATGV-Net:Riegler_2016,MSG-Net:Hui2016,PDSR:Voynov2019,MS-PFL:Chuhua2020,
CA-IRL_SR:Song_2020,RecScanNet:Jeon2018,DDD:Sterzentsenko2019}; commonly, they are trained from paired datasets of low- and high-quality target depth maps.
However, acquiring a large real dataset of this type is challenging and requires a customized, calibrated hardware setup; as a result, such methods commonly rely on downsampling of high-resolution data for constructing training instances; this approach is ineffective for training super-resolution models targeting real depth images heavily contaminated by holes and noise.



One way to circumvent this issue is through the use of \emph{unpaired} learning methods~\cite{
DegradeFirst:Bulat2018_ECCV,Maeda2020,CycleInCycle_SR:Yuan2018}, where the model is trained on two datasets, a dataset of inputs and a dataset of targets, with their elements not necessarily forming pairs.
While, in principle, one can directly apply existing image-based unpaired methods to depth data, these methods tend to significantly underperform in practice as they fail to capture the distinctive characteristics of depth data: 
unlike color photos, depth scans often contain gaps; compared to RGB images where pixels take values from a large but finite palette, the range of depth values is, in principle, continuous and unbounded; additionally, perceptually-relevant differences in depth scans are best captured by depth-specific measures. 


We propose, to the best of our knowledge, the first  learning-based method for depth super-resolution for real-world 
sensor data, using \emph{unpaired} data for training, i.e., a set of raw sensor depth images and a set of high-quality, high-resolution depth images generated by depth fusion, 
without correspondences between images from these sets.

A key ingredient of our method is the introduction of depth enhancement, \ie, a hole-filling and surface denoising method, into the super-resolution pipeline; we demonstrate that coupling depth enhancement and super-resolution tasks yields significant improvements over the baselines.
To implement this, similarly to recent literature (\eg, \cite{DegradeFirst:Bulat2018_ECCV}), we design a two-stage approach for training our method, including (1) an unpaired training stage for a depth degradation model, and (2) a supervised training stage for an enhancement model.


Importantly for evaluating depth super-resolution methods, most existing RGB-D scan datasets cannot be used as they offer either high- or low-resolution sensor depth only. 
To this end, we propose a paired dataset providing both real-world RGB-D scans and high-resolution, high-quality reference depth, that we construct by ray-tracing 3D reconstructions of indoor scenes in ScanNet~\cite{ScanNet:Dai2017} obtained by depth fusion~\cite{BundleFusion:Dai2017}. 
Basing on these, we develop a depth super-resolution and enhancement benchmark, extending a standard evaluation methodology with perceptual measures.




Our  evaluation shows that our method outperforms several state-of-the-art image-to-image translation approaches applied to depth in a pure enhancement mode. 
Likewise our approach outperforms straightforward combinations of deep unpaired enhancement (e.g.,~\cite{Gu2020_TIP}) and bicubic upsampling, emphasizing the need for a close integration of enhancement and super-resolution parts.

To summarize, our contributions are as follows:
\begin{itemize}

\item We introduce the first method designed specifically to perform learning-based, unpaired depth super-resolution, that we call \emph{UDSR}.
Compared to state-of-the-art unpaired depth enhancement techniques adapted to perform SR (\eg, \cite{Gu2020_TIP}), our method shows 28\% improvement in \emph{depth super-resolution performance,} in terms of a perceptual \mse[v] quality measure. 

\item A key component of our method is a new unpaired depth enhancement algorithm that performs efficient unpaired learning, leverages RGB guidance, and optimizes depth-specific performance measures; this results in better denoising and inpainting results for depth images.
Compared to state-of-the-art unpaired depth enhancement~\cite{Gu2020_TIP}, our algorithm achieves 63\% gain in \emph{depth enhancement performance} in terms of the perceptual \mse[v] measure.

\item We present a new benchmark for real-world depth SR and enhancement based on existing datasets~\cite{ScanNet:Dai2017,InteriorNet:Li2018} and a methodology to compare paired and unpaired approaches.

\end{itemize}

\section{Related Work}


\noindent \textit{Depth Super-Resolution (SR).}
Depth super-resolution has been approached from multiple perspectives:  filter-based~\cite{JBU:Kim_2014,JBU:Kopf_2007}, optimization-based~\cite{haefner2019photometric,haefner2018fight}, and data-driven~\cite{ATGV-Net:Riegler_2016, MSG-Net:Hui2016,DJF:Li_2016, PDSR:Voynov2019, MS-PFL:Chuhua2020, CA-IRL_SR:Song_2020, DBPN:Haris_2018, SR_review:Yang_2019, SRFBN:Li_2019,TTSR:Yang_2020, SPSR_GG:Ma_2020}.

In non-learning context, single-frame RGB-guided depth SR has been tackled with joint bilateral filters~\cite{JBU:Kopf_2007} and filters with adaptive smoothing~\cite{JBU:Kim_2014}; optimization-based shape-from-shading approaches~\cite{haefner2019photometric,haefner2018fight} relying on photometric constraints and a number of priors.
While such techniques can likely generalize across sensors, their ability to exploit RGB and depth image-specific characteristics such as distribution of depth values is limited. 

Among data-driven approaches,  CNNs have been used in combination with optimization-based methods~\cite{ATGV-Net:Riegler_2016}, joint filtering methods~\cite{DJF:Li_2016}, as well as  with progressive multi-scale fusion of RGB and depth features~\cite{MSG-Net:Hui2016,MS-PFL:Chuhua2020}; we include an explicit RGB guidance mechanism in our enhancement step, but without applying any optimization to network predictions.
More recently, perceptually-based depth SR~\cite{PDSR:Voynov2019} enabled more accurate surface reconstruction; we integrate their loss function in our training framework.
\cite{CA-IRL_SR:Song_2020} introduced non-linear downsampling degradations to improve robustness of their depth SR method; in contrast, our method automatically captures relevant degradation patterns by a learned depth-to-depth translation step. 
SRFBN~\cite{SRFBN:Li_2019} is an established supervised image SR method often used as a strong baseline for evaluating depth SR; we compare against this approach in our work. The most recent and concurrent work~\cite{he2021towards} is the first which considers depth super-resolution on real sensor data. It is trained on their own collected paired dataset of low- and high-resolution depth maps. However, they rely on image colorization~\cite{ImgColor:Levin_2004} to inpaint holes in input low-resolution depth maps as pre-processing stage. This could be inefficient with big holes and areas of drastic change in depth. Moreover, they do not use depth normals neither for evaluation nor in their approach, which can lead to noisy surfaces reconstructed from their output depth.

Importantly, all these methods need registered pairs of input-output images of the same scene; differently, in this work we explore an \emph{unpaired learning scenario} where the sets of source and target depth images may depict distinct environments. 

\vspace{0.3em}
\noindent \textit{Unpaired Image SR.}
More recently, image-based approaches have resorted to using unpaired learning methods to more accurately model image acquisition and processing artefacts; these formulations bypass the need for paired data, greatly simplifying construction of training datasets. Among unpaired image-based methods, 
Zero-shot SR~\cite{ZSSR:Shocher_2018} relies on self-supervised training from low-resolution images only; other approaches use independent sets of low-resolution and high-resolution images, commonly employing cycle consistency~\cite{CycleGAN:Zhu2017} to learn the SR mapping without correspondences between images.
Most unpaired systems are trained in multiple stages: 
Cycle-in-Cycle~\cite{CycleInCycle_SR:Yuan2018} learns image cleaning during the first cycle and SR in the unpaired setting using the second cycle,
Bulat~\etal~\cite{DegradeFirst:Bulat2018_ECCV} learns degradation using an unpaired setup, further performing supervised training for SR.
Maeda~\cite{Maeda2020} decomposes SR mapping into a cleaning step trained in an unpaired way and a pseudo-supervised SR network; similarly, our method uses two-stage training.

\vspace{0.3em}
\noindent \textit{Depth Enhancement~\cite{Gu2020_TIP,RecScanNet:Jeon2018}} 
is an umbrella task that encompasses denoising~\cite{DnCNN:Zhang2017, NL_Denoising:Lefkimmiatis_2017, UDN:Lefkimmiatis_2018, CBDNet:Guo_2019}, completion, and inpainting~\cite{DC:Zhang2018, PConvlInpainting:Liu2018, CAInpainting:Yu_2018, StructureFlow:Ren2019, HiFIll:Yi_2020}. 
Among the methods in this group, our approach implements a data generation technique similar to LapDEN~\cite{RecScanNet:Jeon2018} who render 3D reconstructions of scenes in ScanNet~\cite{ScanNet:Dai2017} to obtain training data for their depth enhancement method; we also compare our method to this approach.

\vspace{0.3em}
\noindent \textit{Unpaired Depth Enhancement} is similar in spirit to unpaired image-to-image translation but requires considerable adaptation of existing image-based methods.  
Gu~\etal pioneered a GAN-based unpaired depth enhancement~\cite{Gu2020_TIP} with a four-stage learnable approach, involving hole prediction, image adaptation, degradation, and final enhancement.
From a self-supervised perspective, \cite{DDD:Sterzentsenko2019} leverages photometric constraints to recover high-quality depth but requires a non-standard acquisition setup.
Learning from unsynchronized low- and high-quality depth frames,~\cite{Shabanov_2020_3DV} proposes a self-supervised approach employing temporal and spatial alignment.
Among these methods, we extensively compare to single-image unpaired enhancement~\cite{Gu2020_TIP}, adapt this method for depth SR via complementing it with various upsampling methods.

\vspace{0.3em}
\noindent \textit{RGB-D Datasets for Depth SR.}
Among RGB-D datasets, Middlebury~\cite{Middlebury14:Scharstein2014}, NYU-Depth V2~\cite{NYUv2:Silberman2012}, SUN RGB-D~\cite{SUNRGBD:Song_2015}, and the synthetic ICL-NUIM~\cite{ICL-NUIM:Handa_2014} provide RGB-D frames but cannot serve as evaluation data for depth super-resolution since they either do not provide real-world sensor depth or lack corresponding ground truth.
Matterport3D~\cite{Matterport3D:Chang_2017} is a large-scale dataset with high-quality depth but lacks corresponding depth from less accurate sensors.
ToF-Mark~\cite{ToF-Mark:Ferstl_2013} contains depth maps from a low-accuracy time-of-flight sensor and a high-accuracy structured light scanner but only provides three frames. 
Similarly, Redwood~\cite{Redwood:Park2017} consists of RGB-D sequences obtained using a consumer Asus Xtion Live depth camera and point clouds from industrial-grade laser scanner but captures only five scenes.

In the context of depth enhancement, \cite{RecScanNet:Jeon2018} synthesized a paired dataset from ScanNet~\cite{ScanNet:Dai2017}, a large-scale collection of RGB-D scans, using its complete 3D reconstructed models obtained using  BundleFusion~\cite{BundleFusion:Dai2017}.
These models were ray-casted to obtain image pairs of the same resolution for training neural networks targeting denoising and hole-filling.
We extend this approach to our task, additionally creating high-resolution depth images from renderings of 3D reconstructions (Section~\ref{sec:benchmark}). 
We additionally conduct experiments using synthetic indoor data in InteriorNet~\cite{InteriorNet:Li2018}.

\section{Unpaired Depth Super-Resolution Framework}
\label{sec:method}

\subsection{Unpaired Depth Super-Resolution}
\label{method:udsr}

\noindent \emph{Task Formulation.}
RGB-D sensors capture the surface of the scene using pairs~$(\image, \dmapSrc)$ of sensor data, where $\image$ is a color image and~$\dmapSrc$ a depth image, acquired jointly.
Following the existing practice in depth SR literature, we assume color images to come at high resolution and satisfactory quality; differently to color photos, depth images commonly suffer from low resolution, high noise, and incomplete depth values.
These limitations motivate the \emph{depth super-resolution} (depth SR) task, aimed at constructing an improved depth image~$\dmapTgt$ given the sensor data $(\image, \dmapSrc)$.

A common approach to depth SR is to fit a mapping from sensed instances~$(\image, \dmapSrc)$ to desired depth~$\dmapTgt$ using a neural network~\cite{ATGV-Net:Riegler_2016, MSG-Net:Hui2016,DJF:Li_2016, PDSR:Voynov2019, MS-PFL:Chuhua2020, CA-IRL_SR:Song_2020, DBPN:Haris_2018, SR_review:Yang_2019, SRFBN:Li_2019,TTSR:Yang_2020, SPSR_GG:Ma_2020}.
In this paper, we refer to this learning-based formulation as \emph{supervised (paired)} depth SR.
During training, a supervised learning algorithm must query reference depth values in densely sampled surface points where predictions will be made; hence, it requires \emph{pairs} of source and target RGB-D or depth images of the scene that are \emph{registered} (\ie, brought to a shared reference frame). 
Synthetic scenes can be sampled at arbitrary resolution to provide such data; however, models trained on purely synthetic data struggle to generalize to real images. 
On the other hand, acquiring real-world data using multiple depth sensors remains challenging as one needs to perform precise alignment of captured data, \eg, by constructing a custom hardware setup.
As a result, training data are commonly \emph{synthesized} by simply downsampling clean, high-resolution images; doing this leaves out modelling acquisition artefacts such as noise and gap patterns occurring in real-world images (a challenging task in itself~\cite{berger2013benchmark,ley2016syb3r}).
This mismatch in distributions of training and evaluation data  becomes evident as models trained using synthetic downsampling fail to produce fully denoised, complete outputs, as shown in Figure~\ref{fig:sr_fails}.

As an alternative formulation, in this work we consider \emph{unpaired} depth SR, a previously unexplored task that does not rely on having paired image data during training; instead, our formulation assumes that source and target image datasets are entirely independent.
In practice, this means that sensor data pairs~$(\image^{\src}, \dmapSrc)$ and target instances~$(\image^{\tgt}, \dmapTgt)$ can capture non-overlapping segments of the scene or even separate scenes; they may be collected using differing, non-registered sensors, or at different periods of time (\eg, allowing the scenes to undergo changes).
Such data is significantly easier to collect, as no alignment or additional hardware are necessary, and facilitates re-use of existing datasets; as a consequence, our formulation enables a broader scope of applications.

\vspace{0.5em}

\noindent \emph{Framework Overview.}
The input to our algorithm is a single RGB-D image with a low-resolution, noisy, incomplete depth; as an output, we produce a single high-resolution, denoised, complete depth image.
To achieve our objective, we designed a learning-based framework to learn depth SR without paired, registered training instances; we made a series of algorithmic decisions to address the challenges of our task.
The four main components of our method are:
\begin{enumerate}

\item \emph{Framework Achitecture} (Section~\ref{method:architecture}). 
We follow a two-step approach to depth super-resolution: first, we upsample the input depth image to the desired output resolution; second, we process the upsampled depth image using our enhancement algorithm to produce the final result.
In our system, only the enhancement part is trained by solving a two-stage unpaired learning task.

\item \emph{Enhancement Algorithm} (Section~\ref{method:enhancement}). 
We pre-train four neural sub-networks using complementary sub-tasks to provide rich photometric and geometric features, and integrate these models within our learning-based enhancement algorithm~$\openh$.
Our best performing learning configuration predicts an enhanced depth image from the upsampled input RGB-D image, its feature maps, and an intermediate depth estimate.

\item \emph{Unpaired Translation Algorithm} (Section~\ref{method:translation}). 
We construct supervised data for training our enhancement algorithm by synthesizing a realistic \textit{pseudo-source} RGB-D instance for each \textit{target} (high-resolution) RGB-D image.
To this end, we pre-train a deep translation network~$\GenTgtSrc$ using an unpaired, cycle-consistent adversarial learning approach, minimizing discrepancy between sets of source and target instances.

\item \emph{Construction of Datasets} (Section~\ref{sec:benchmark}). 
We develop a methodology to directly evaluate paired and unpaired depth SR algorithms and construct three benchmarks  of up to 38,000 instances for our quantitative comparisons.

\end{enumerate}

In the next sections, we describe each component in detail and provide a rationale for our algorithmic choices.

\begingroup
\renewcommand{\arraystretch}{1.3}
\begin{table}[t]
\centering
\begin{tabular}{l r r r }
\textbf{Symbol} & \multicolumn{2}{c}{\textbf{Depth Properties}}  &\textbf{Set Meaning}  \\ 
\cline{2-3}
 & \textbf{Resolution} & \textbf{Quality} &  \\ \hline
$\src$          & $w \times h$      & low   & Source RGB-D images \\
$\srcup$        & $kw \times kh$    & low   & $\src$ with upsampled depth \\
$\tgt$          & $kw \times kh$    & high  & Target RGB-D images \\
$\tgtdown$      & $w \times h$      & high  & $\tgt$ with downsampled depth \\
$\pseudosrc$      & $w \times h$      & $\approx$low  & Pseudo-source RGB-D images \\
\hline
\end{tabular}%
\caption{Overview of the qualitatively different sets of RGB-D images $(I, D)$ used within the present work.
\label{table:symbols}
\enquote{Low-quality} depth images are prone to noise and incomplete depth values; \enquote{high-quality} depth images are clean and complete.}
\end{table}
\endgroup

\subsection{Task Decomposition and Framework Architecture}
\label{method:architecture}

\begingroup
Our depth SR algorithm is designed to increase the spatial resolution (\ie, do upsampling) and to suppress noise, fill in missing areas, and resolve detail (\ie, do enhancement) in an input depth image.
We have considered a number of alternatives for implementing these diverse modifications in the unpaired learning context. 
We summarize our resulting configuration and compare it to alternatives below,
provide more detail in Sections~\ref{method:enhancement} and~\ref{method:translation}, and discuss the effect of various choices in Section~\ref{sec:experiments}.

\setlength{\intextsep}{0pt}%
\begin{wrapfigure}{r}{0.4\columnwidth}
\begin{center}
\includegraphics[width=0.4\columnwidth]{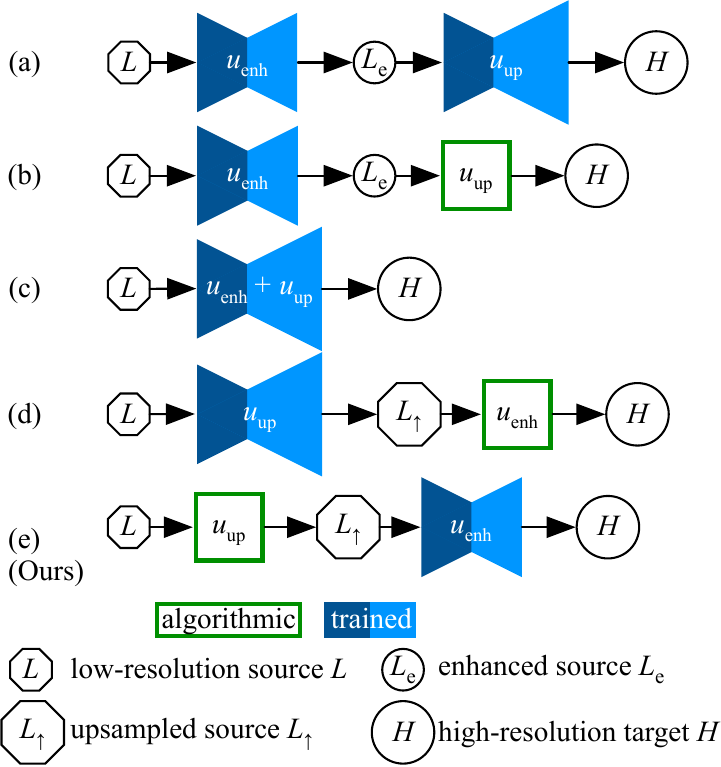}
\end{center}
\vspace*{-3mm}
\caption{Architectural options we considered (see also Table~\ref{table:symbols} for notation).
Our architectural choice (e) enjoys both a simple design and high performance in experiments.
\textit{Enhanced source} $\src_{\text{e}}$ refers to a set of RGB-D images with low-resolution but enhanced depth (used only in this scheme).}
\label{fig:decomposition_variants}
\end{wrapfigure}

\vspace{0.5em}

\noindent \emph{Task Decomposition.} 
We decompose the depth SR mapping into a sequence $\opsr = \openh \circ \opup$ where a \textit{trainable} enhancement algorithm~$\openh$ succeeds (an optional) \textit{non-trainable} upsampling operation~$\opup$ (see Figure~\ref{fig:decomposition_variants}~(e)); separating these stages is in line with recent successful approaches to unpaired image SR~\cite{DegradeFirst:Bulat2018_ECCV}.
Compared to complementing a \textit{trainable} enhancement algorithm with a separate \textit{trainable} upsampling network (\cf~\cite{DegradeFirst:Bulat2018_ECCV}, Figure~\ref{fig:decomposition_variants}~(a)), our approach requires training only a single model; adding a \textit{non-trainable} upsampling operation (Figure~\ref{fig:decomposition_variants}~(b)) yields results inferior to our algorithm, according to our experiments.
In comparison to integrating the enhancement and upsampling stages inside a \textit{single trained} network (Figure~\ref{fig:decomposition_variants}~(c)), or complementing a \textit{trained} upsampling network with a \textit{non-trained} enhancement operation (Figure~\ref{fig:decomposition_variants}~(d)), our approach enjoys greater simplicity by avoiding the need for in-network upsampling~\cite{Maeda2020}.
Our upsampling operation~$\opup$ is bicubic interpolation: we upscale the input $w \times h$ depth image to the output $kw \times kh$ resolution ($k$ being the SR factor). The enhancement algorithm~$\openh$ takes in an RGB-D image with the upsampled depth and produces a refined depth image at the same spatial resolution.
Making upsampling optional in this way enables using our algorithm at different SR factors, including keeping the input resolution of the image ($k = 1$).

\endgroup

\vspace{0.5em}

\begingroup

\begin{figure}[t]
\centering
\includegraphics[width=0.7\columnwidth]{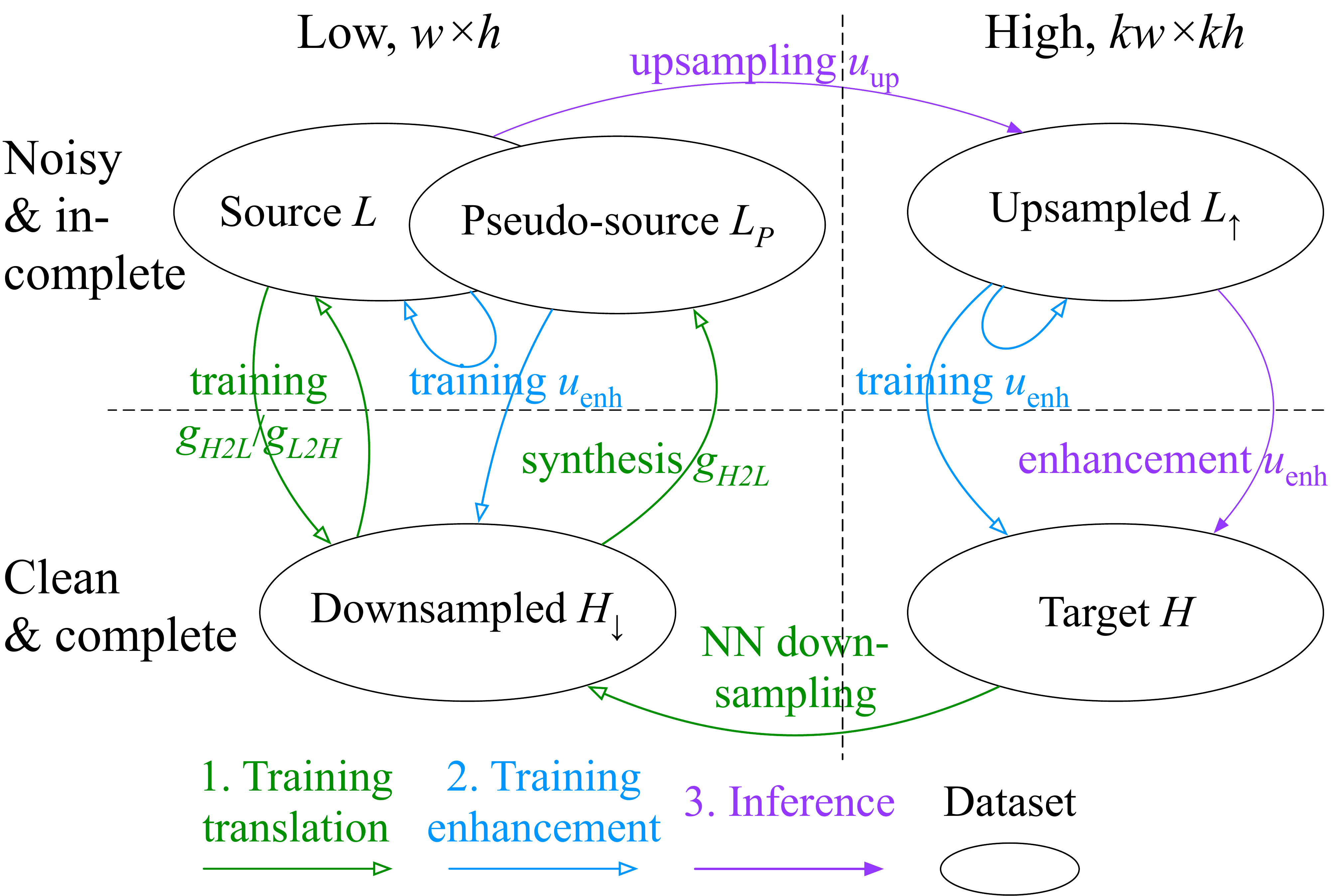}
\caption{Overview of the conceptual architecture and principal stages of our learning framework. 
(1) We start with the \textcolor{ForestGreen}{training of the unpaired translation algorithm} and obtain bidirectional mappings $\GenTgtSrc, \GenSrcTgt$ between the source $\src$ and the downsampled $\tgtdown$ sets. 
In the same stage, we \textcolor{ForestGreen}{synthesize} the pseudo-source $\pseudosrc$ set by translating instances in $\tgtdown$.
(2) Next, we \textcolor[rgb]{0,0.5,1}{train the enhancement algorithm $\openh$} at low and high resolution using a multi-task objective.
(3) During inference for a source instance from $\src$, we perform \textcolor[rgb]{0.58,0.22,1}{upsampling by $\opup$} and \textcolor[rgb]{0.58,0.22,1}{enhancement by $\openh$} to compute the final prediction.}
\label{fig:fig_datasets}
\end{figure}

\endgroup

\noindent \emph{Achitecture of the Learning Framework.} 
Conceptually, we reframe our unpaired learning task as two arguably easier tasks: a data synthesis task, concerned with \emph{generating appropriate training (pseudo-source) images,} and a supervised regression learning task, which addresses \emph{learning an enhancement mapping from the pseudo-sources} and bears similarity to recent successful approaches to image SR~\cite{Maeda2020}.
We elaborate on the design of our framework below; we visually accompany this design by Figure~\ref{fig:fig_datasets} and present an overview for the symbols denoting the different sets of RGB-D data used throughout this work in Table~\ref{table:symbols}. 

The input domain for our depth SR method is the set of source RGB-D images with low-resolution depth \textit{(source~$\src$),} that we upsample using~$\opup$ to obtain RGB-D images with upsampled depth \textit{(upsampled~$\srcup$)}.
Our central learning goal is to construct an enhancement algorithm~$\openh$ that transforms the upsampled~$\srcup$ to the set of target RGB-D images with high-resolution depth \textit{(target~$\tgt$)}.
In this context, our first objective is to construct a collection of training RGB-D instances with source-like and target-like depth, suitable for supervised training of the enhancement algorithm.
To serve as such source-like and target-like images, we define two sets with $w \times h$ depth: pseudo-source RGB-D images \emph{(pseudo-source~$\pseudosrc$)} and target RGB-D images with downsampled depth \textit{(downsampled~$\tgtdown$)}.
The downsampled~$\tgtdown$ is obtained by nearest-neighbour downsampling of the depth images in the target set~$\tgt$. 
Pseudo-sources~$\pseudosrc$ are constructed by training an unpaired image-to-image translation method minimizing discrepancy between the distributions of the pseudo-sources~$\pseudosrc$ and the sources~$\src$. 
Our second objective is to use the paired data in the pseudo-source~$\pseudosrc$ and downsampled~$\tgtdown$ sets to learn an effective enhancement method.
During training, we strive to maintain equally high enhancement performance for the pseudo-source~$\pseudosrc$, source~$\src$, and upsampled~$\srcup$ sets despite differences in their distributions and resolution.

Following this design, we decompose our learning framework into two interrelated trainable parts: the unpaired translation algorithm, implementing the bidirectional translation between the downsampled~$\tgtdown$ and the source~$\src$ sets, and the enhancement algorithm, implementing the enhancement transformation~$\openh$.
We train them consecutively using two stages. 
In the first stage, we train the translation algorithm using the source~$\src$ and the downsampled~$\tgtdown$ in an unpaired manner, obtaining a deep translation network~$\GenTgtSrc: \tgtdown \to \src$ (Section~\ref{method:translation}), and freeze the weights of~$\GenTgtSrc$.
In the second stage, we use $\GenTgtSrc$ to process samples in the downscaled~$\tgtdown$ and obtain the set of pseudo-sources~$\pseudosrc$. 
We train the enhancement algorithm using the synthesized pseudo-sources~$\pseudosrc$, sources~$\src$, and downsampled targets~$\tgtdown$ (Section \ref{method:enhancement}), and fine-tune it using the upsampled sources~$\srcup$ and the targets~$\tgt$.
We give details of these components below.

\subsection{Enhancement Algorithm}
\label{method:enhancement}

\begin{figure*}[t]
\centering
\includegraphics[width=\textwidth]{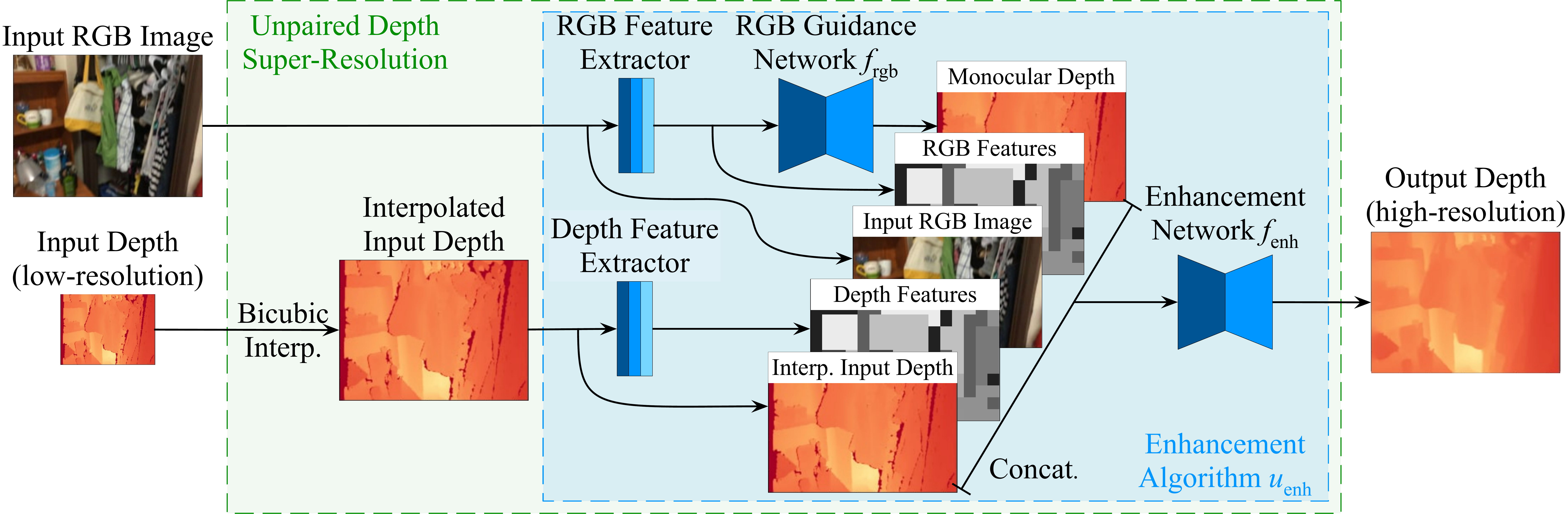}
\caption{Scheme of depth super-resolution with our enhancement algorithm $\openh$. We first upsample depth via Bicubic interpolation to the target resolution of the color image. Then we employ RGB guidance  network $\fimg$ to extract a monocular depth estimate and run feature extractors to obtain photometric and geometric convolutional features. Finally, the enhancement network $\fenh$ predicts a refined depth image by fusing this information along with the source RGB-D pair.}
\label{fig:fig_inference_enhancement}
\end{figure*}

Our enhancement algorithm~$\openh$ is designed to leverage multiple complementary sensed and predicted images as presented in a data-flow Figure~\ref{fig:fig_inference_enhancement}.
More specifically, we include raw input color and depth images, estimate an accurate intermediate monocular depth image from the input color image, and construct two additional photometric and geometric feature maps extracted from the input color and depth images, respectively.
During training, we (1) pre-train a separate RGB guidance network (feature extractor) using RGB-D data with low-resolution depth: source~$\src$ and downsampled~$\tgtdown$; (2) train an enhancement network using a mix of source~$\src$, pseudo-source~$\pseudosrc$ and target~$\tgt$ sets in a shared training loop. 
For inference with an input RGB-D image, we 
(1) upsample its input depth to the target resolution ($kw \times kh$), 
(2) compute feature maps of the input RGB and depth images using convolutional feature extractors,
(3) estimate an intermediate depth image from the RGB image with the RGB guidance network,
and (4) estimate the final depth image from the input RGB-D image concatenated with their feature maps and the intermediate depth image, using the enhancement network.

We describe the supervised training procedure for our enhancement algorithm below; we provide details on the construction of its training data in Section~\ref{method:translation}.
We assume that we have access to RGB-D images from the source~$\src$, pseudo-source~$\pseudosrc$, target~$\tgt$, and downsampled~$\tgtdown$ sets.
We obtain depth images in the pseudo-source~$\pseudosrc$ by applying the translation network (Section~\ref{method:translation}) to those in the target~$\tgtdown$; depth images in the downsampled~$\tgtdown$ are obtained by nearest-neighbour downsampling of the depth from the target~$\tgt$. 
The distributions of the source~$\src$ and pseudo-source~$\pseudosrc$ sets are assumed to be close to identical.

\vspace{0.5em}
\noindent \textit{RGB Guidance Network.}
Color guidance has been established as an important visual cue heavily utilised for RGBD-based depth SR~\cite{MSG-Net:Hui2016} due to the strong correlation between color and depth images. 
Following this intuition, we exploit a simple color guidance mechanism, that we demonstrate to serve as an effective signal, particularly for recovering accurate depth in regions where sensor depth measurements are missing or unreliable.
To achieve this, we define an RGB guidance CNN~$\fimg$ used to estimate a depth image from the respective monocular color image.
We implement~$\fimg$ using a U-Net-like~\cite{UNet:Ronneberger2015} encoder-decoder architecture with two encoders, one for the source~$\src$ and one for the downsampled~$\tgtdown$ sets, and a shared decoder. 
We pre-train~$\fimg$ by minimizing mean absolute error between the monocular depth estimate~$\fimg(\image)$ and the reference depth image $\dmap$ for a given RGB-D image~$(\image, \dmap) \in \src$ or~$(\image, \dmap) \in \tgtdown$, and freeze its weights after pre-training.
We describe the architecture used for~$\fimg$ in the experimental Section~\ref{experiments:setup}.


\vspace{0.5em}
\noindent \textit{Feature Extractors.}
We separately mention photometric and geometric feature extractors, represented by two smaller CNNs trained jointly with the RGB guidance CNN and the enhancement CNN, respectively.
Their outputs, photometric~$x_{\image}$ and geometric~$x_{\dmap}$ convolutional features, are extracted from the raw input RGB and depth images, respectively, and used as additional inputs to our enhancement algorithm (see Figure~\ref{fig:fig_inference_enhancement}).

\vspace{0.5em}
\noindent \textit{Enhancement Network~$\fenh$.}
We define an enhancement CNN~$\fenh$ as our estimator of the final high-resolution depth image~$\dmapTgtEst$ using the diverse available  visual data. 
For an input RGB-D image~$(\image, \dmap)$, we compute a depth estimate~$\fimg(\image)$ using the pre-trained image guidance network~$\fimg$ and extract photometric and geometric convolutional features $x_{\image}, x_{\dmap}$, respectively.
The 5-tuple $(\image, \dmap, \fimg(\image), x_{\image}, x_{\dmap})$ is fed in as an input data for predicting the output~$\dmapTgtEst$ using~$\fenh$ (see Figure~\ref{fig:fig_inference_enhancement}).
The network architecture used for~$\fenh$ is specified in Section~\ref{experiments:setup}.

\vspace{0.5em}
\noindent \textit{Loss Terms for Training~$\fenh$.}
In our training scenario, the network is unlikely to automatically pick up relevant patterns in absense of \enquote{true} large-scale paired data; we thus seek to give our regression model the desired properties by optimizing it using a combination of loss terms, that we find to directly impact performance for our algorithm.
We describe these properties and the respective loss terms here.

We define pixel-weighted mean absolute error (\mae) and mean squared error (\mse) by computing integral of the pixelwise deviation in the predicted and reference depth images
\begin{align}
\label{eqn:depth_based_loss}
&L_p(\dmap_1, \dmap_2) = \norm{w_{p} \odot (\dmap_1 - \dmap_2)}_p, \\
&\mae = L_{1},
\quad
\mse = L_{2}, \nonumber
\end{align}
where $\odot$ denotes pixel-wise multiplication, and use a combination of these with our \textit{depth-based loss term}
\begin{align}
\label{eqn:depth_based_loss_sum}
\LossDepth &= \lambda_{\text{depth}, 1} \mae + \lambda_{\text{depth}, 2} \mse \\
&=
\begin{bmatrix}
\lambda_{\text{depth}, 1} & \lambda_{\text{depth}, 2}
\end{bmatrix}
\begin{bmatrix}
\mae & \mse
\end{bmatrix}^{\intercal}
 = \bm{\lambda}_{\text{depth}} \bm{L}_{\text{depth}}^{\intercal} \nonumber
\end{align}
where $\bm{\lambda}_{\text{depth}}$ weights $p$-norms of pixewise deviations. 

While the two depth images may be close in the Euclidean sense, appearance of their respective surfaces (as captured, \eg, by a rendering) can vary significantly due to geometric noise in local surface orientation. 
Thus, we assess surface quality using perceptual losses~\cite{PDSR:Voynov2019} defined as $p$-norms of pixelwise-weighted difference in surface renderings of depth images averaged over three orthogonal light directions~$\lightDir_i$: 
\begin{align}
\label{eqn:actual_mse_v}
&R_p(\dmap_1, \dmap_2) = \frac{1}{3} \sum_{i=1}^3 \norm{w_{p} \odot (\nmap_1 - \nmap_2) \cdot \lightDir_i}_p, \\
&\mae[v] = R_{1},
\quad
\mse[v] = R_{2}, \nonumber
\end{align}
where $\nmap_1$ and $\nmap_2$ are finite-difference estimates of per-pixel normals from the depth images $\dmap_1$ and $\dmap_2$, respectively.
We define our \textit{surface-based loss term} via
\begin{align}
\label{eqn:mse_v}
\LossPerceptual 
&= \lambda_{1} \mae[v] + \lambda_{2} \mse[v] \\
&=
\begin{bmatrix}
\lambda_{\text{surf}, 1} & \lambda_{\text{surf}, 2}
\end{bmatrix}
\begin{bmatrix}
\mae[v] & \mse[v]
\end{bmatrix}^{\intercal}
 = \bm{\lambda}_{\text{surf}} \bm{L}_{\text{surf}}^{\intercal} \nonumber
\end{align}

We specify our pixelwise weighting functions $w(u), u = (i, j)$ for use within the depth- and surface-based loss terms \eqref{eqn:depth_based_loss}, \eqref{eqn:mse_v} according to the general expression
\begin{equation}
\label{eq:pixelwise_weighting_functions}
w(u) = w_1 \chi_{A_1}(u) + w_1 \chi_{A_2}(u)
 = \begin{bmatrix}w_1 & w_2\end{bmatrix} 
 \begin{bmatrix}\chi_{A_1}(u) & \chi_{A_2}(u)\end{bmatrix}^{\intercal}
 = \bm{w} \bm{\chi}(u), 
\end{equation}
where $w_1, w_2 \in \mathbb{R}$, the function $\chi_{A_1}(\cdot)$ is an indicator function of the set $A_1$ of \textit{valid} pixels, and $\chi_{A_2}(\cdot)$ is an indicator function of the set $A_1$ of pixels corresponding to \textit{gaps}.
As we would like to plausibly inpaint gaps, we typically set $w_2$ to a larger value compared to $w_1$. 
We specify particular values for per-pixel weights $w_p$ in Section~\ref{sec:experiments}.

To encourage sharp depth discontinuities at object boundaries, we use the \textit{edge-based regularizer term}~\cite{T2Net:Zheng_2018}
\begin{equation}
\RegEdge (\image, \dmap) 
 = \norm{\nabla_h \dmap }_1e^{-\norm{\nabla_h \image}_1}
 + \norm{\nabla_v \dmap }_1e^{-\norm{\nabla_v \image}_1},
\end{equation}
where $\nabla_h$ and $\nabla_v$ denote finite differences computed in horizontal and vertical direction.

To aid noise suppression and impose spatial smoothness on the normals~$\nmap$ for an RGB-D image $(\image, D)$, we additionally optimize the \textit{total variation regularizer term}~\cite{Primal-Dual:Riegler_2016,CycleInCycle_SR:Yuan2018} expressed by 
\begin{equation}
\RegSmooth (\nmap)
 = \norm{\nabla_h \nmap}_2
 + \norm{\nabla_v \nmap}_2.
\end{equation}

\noindent \textit{Training Procedure for~$\fenh$.}
Our core idea is to implement a shared enhancement mapping for both source real-world~$\src$ and pseudo-source synthesized~$\pseudosrc$ sets using a single model, enabling it to treat these sets in a unified way; this leads us to include two critical ingredients in designing our procedure for training the enhancement network~$\fenh$.
First, to effectively perform enhancement we require the network to minimize a \textit{pseudo-supervised objective,} aiming to reconstruct output depth in~$\tgtdown$ from respective synthesized images in~$\pseudosrc$.
For pseudo-source samples in $\pseudosrc$, as accurate reference depth and normals are available, we minimize 
\begin{equation}
\Loss^{\pseudosrc}_{\text{enh}} (\fenh)
 = \LossDepth(\dmap^{\pseudosrc}, \widehat{\dmap}^{\tgtdown})
 + \LossPerceptual(\dmap^{\pseudosrc}, \widehat{\dmap}^{\tgtdown})
 + \lambda^{\pseudosrc}_{\smooth} \RegSmooth(\widehat{\dmap}^{\tgtdown}),
\label{eq:enh_pseudosrc_objective}
\end{equation}
where $\widehat{\dmap}^{\tgtdown} = \fenh(\image, \dmap^{\pseudosrc})$ is the refined depth image produced by the enhancement CNN.
However, this alone does not guarantee achieving similar performance for real-world images in~$\src$ due to a discrepancy between $\src$ and $\pseudosrc$. 
We thus additionally minimize a \textit{self-supervised objective} over instances from~$\src$ which serve as a fixed point constraint.
%
For these instances, normals tend to be very noisy, and we thus exclude the surface-based term $\LossPerceptual$ and instead focus on detecting object contours using $\RegEdge$ by minimizing
\begin{equation}
\Loss^{\src}_{\text{enh}} (\fenh)
 = \LossDepth(\dmap^{\src}, \widehat{\dmap}^{\src})
 + \lambda_{\edge}^{\src} \RegEdge (\widehat{\dmap}^{\src})
 + \lambda_{\smooth}^{\src} \RegSmooth (\widehat{\dmap}^{\src}),
\label{eq:enh_src_objective}
\end{equation}
where we expect the output depth image $\widehat{\dmap}^{\src} = \fenh(\image, \dmap^{\src})$ to replicate the input depth $\dmap^{\src}$.

Note that each of the terms $\LossDepth(\dmap^{\pseudosrc}, \widehat{\dmap}^{\tgtdown})$, $\LossPerceptual(\dmap^{\pseudosrc}, \widehat{\dmap}^{\tgtdown})$, and $\LossDepth(\dmap^{\src}, \widehat{\dmap}^{\src})$ incorporate 1-norm and 2-norms weighted by respective weights $\bm{\lambda}^{\pseudosrc}_{\text{depth}}$, $\bm{\lambda}^{\pseudosrc}_{\text{surf}}$, and $\bm{\lambda}^{\src}_{\text{depth}}$.

Our full enhancement objective is
\begin{equation}
\Loss_{\text{enh}}
 = \Loss^{\pseudosrc}_{\text{enh}}
 + \Loss^{\src}_{\text{enh}}
 \label{eq:enh_full_objective}
\end{equation}
where $\Loss^{\pseudosrc}_{\text{enh}}$ leverages the available pseudo-supervision and $\Loss^{\src}_{\text{enh}}$ helps to achieve good performance for real instances.
Technically, we include an equal number of samples from both $\pseudosrc$ and $\src$ in each mini-batch while performing gradient descent.

\noindent \textit{Training at High Resolution.}
We pre-train the enhancement network at $w \times h$ resolution using instances in~$\pseudosrc$, $\tgtdown$, and $\src$; after that, we fine-tune it using $kw \times kh$ high-resolution images. 
We keep the same objective in~\eqref{eq:enh_pseudosrc_objective}--\eqref{eq:enh_full_objective} but set new hyperparameters giving heavier weight to depth and surface terms.
Section~\ref{sec:experiments} describes specific values for weighting, details of our parameter choices, and results of an ablation study involving alternatives.




\subsection{Unpaired Translation Algorithm}
\label{method:translation}

We design our translation step to perform data-driven construction of pseudo-sources~$\pseudosrc$ required for training the enhancement algorithm~$\openh$ from the downsampled targets~$\tgtdown$.
To this end, we take inspiration from the cycle-consistent learning paradigm~\cite{CycleGAN:Zhu2017} and its recent applications to photo and depth SR~\cite{DegradeFirst:Bulat2018_ECCV,Maeda2020,CycleInCycle_SR:Yuan2018,Gu2020_TIP}, and construct two deep models for performing forward and reverse translation between the downsampled~$\tgtdown$ and the source~$\src$.
Note that performing this training under the cycle-consistent adversarial framework does not require having paired training instances as such models minimize objectives formulated in terms of distributions.
Once the forward translation network has been trained, we may proceed with generation of pseudo-source instances~$\pseudosrc$ via passing through it images from the downsampled~$\tgtdown$.

\vspace{0.5em}
\noindent \emph{Translation Networks.}
We define a forward translation network~$\GenTgtSrc$ to learn a desired \textit{degradation mapping} from the high-quality downsampled~$\tgtdown$ to the low-quality source~$\src$, and a concurrent reverse translation network~$\GenSrcTgt$ which learns to translate the source~$\src$ to the downsampled~$\tgtdown$. 
Depth images in both sets have the same $w \times h$ spatial resolution.
Operation of these networks can be described in terms of relations 
\begin{equation}
\widehat{\dmap}^{\src} = \GenTgtSrc(\image, \dmap^{\tgtdown})
\quad
\text{and}
\quad
\widehat{\dmap}^{\tgtdown} = \GenSrcTgt(\image, \dmap^{\src}),
\label{eq:translation_nets}
\end{equation}
where $\widehat{\dmap}^{\src}$ is the translated pseudo-source depth for a downsampled~$(\image, \dmap^{\tgtdown}) \in \tgtdown$, and $\widehat{\dmap}^{\tgtdown}$ is the translated downsampled depth of a source RGB-D image~$(\image, \dmap^{\src}) \in \src$. 
We additionally define a reconstructed depth image $\dmapTgtdownRec$ via 
\begin{equation}
\dmapTgtdownRec = \GenSrcTgt(\image, \GenTgtSrc(\image, \dmap^{\tgtdown})).
\label{eq:cycle_consistency}
\end{equation}

\begin{figure}[t]
\centering
\includegraphics[width=0.6\columnwidth]{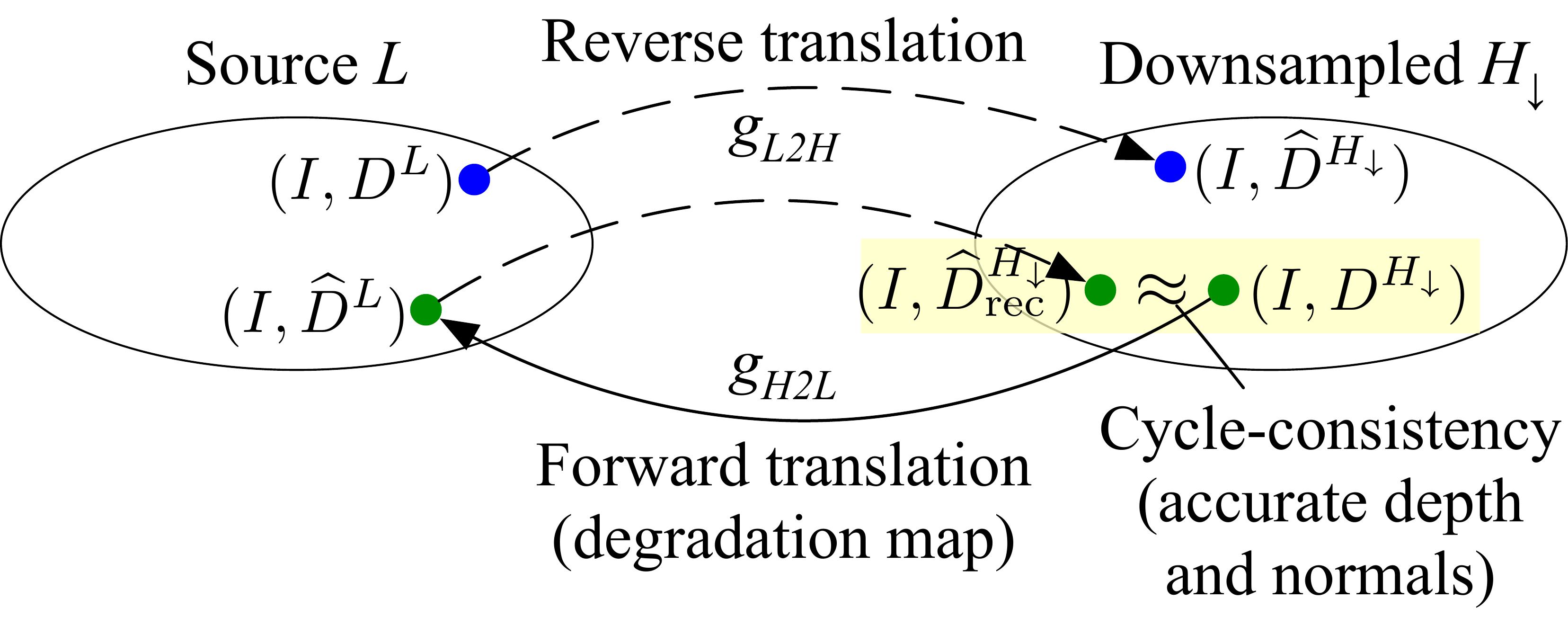}
\caption{We use the source~$\src$ and the downsampled~$\tgtdown$ as the two sets of RBG-D images involved in unpaired training of the translation network.}
\label{fig:datasets_translation}
\end{figure}


\vspace{0.5em}
\noindent \emph{Objective for Unpaired Training.}
We train the forward and reverse translation networks using several distinct loss terms. 
We formulate an \textit{adversarial loss term} by adapting an existing image-to-image translation objective developed in the context of image SR~\cite{Maeda2020}. 
While a straightforward adaptation already allows to optimize translation networks (generators)~$\GenSrcTgt$ and~$\GenTgtSrc$ along with discriminators~$\discDmapTgt$ and~$\discDmapSrc$, to further raise performance for RGB-D data we develop an improved formulation by integrating three modifications (we investigate their effect in Section~\ref{sec:experiments}). 
First, in a similar vein to~\cite{Maeda2020}, we substitute the original~$\dmap^{\tgtdown}$ with a reconstructed~$\dmapTgtdownRec$ for the discriminator~$\discDmapTgt$ operating in the downsampled~$\tgtdown$: we aim to relax the requirements for the generator and to stabilize training.
Second, to facilitate reconstruction of faithful normals, we extend the adversarial part with surface normals discriminator networks~$\discNmapTgt$ and~$\discNmapSrc$ operating on images in the downsampled~$\tgtdown$ and the source~$\src$, respectively (we compute normals using finite differences).
Finally, we formulate all adversarial losses in terms of the Least Squares GAN~\cite{LSGAN:Mao2017} that is known to improve GAN performance by preventing gradient saturation via a least squares penalty. 
We provide the full formulation of our adversarial loss term~$\Loss_{\text{adv}}$ in the Supplementary material. 
Figure~\ref{fig:fig_translation_component_training} illustrates the data flow used in during training of our translation network. 

By the cycle-consistency property, we expect the reconstructed depth image $\dmapTgtdownRec$ from~\eqref{eq:cycle_consistency} to approximate the original image well: $\dmapTgtdownRec \approx \dmap^{\tgtdown}$. 
To implement cycle-consistent training, we enforce the accurate reconstruction of the original depth image when undergoing a composition of forward and reverse mapping by minimizing the \textit{cycle-consistency term} (see Figure~\ref{fig:datasets_translation})
\begin{equation}
\Loss_{\text{cycle}} = \mae + \mse[v],
\end{equation}
emphasizing the need to fit both depth and normals components accurately ($\mae$ and $\mse[v]$ are defined in~\eqref{eqn:depth_based_loss} and~\eqref{eqn:depth_based_loss_sum}, respectively).
As the downsampled~$\tgtdown$ is the only set that provides accurate reference depth and normals, we compute $\Loss_{\text{cycle}}(\dmapTgtdownRec, \dmap^{\tgtdown})$ for instances in $\tgtdown$ only.

The source~$\src$ and downsampled~$\tgtdown$ sets have different range and distribution of depth values; to prevent both networks~$\GenSrcTgt$ and~$\GenTgtSrc$ from learning systematic shifts, we regularize translation via the two \textit{range regularization terms}
\begin{align}
\Reg^{\src}_{\text{range}}(\dmapSrc) &= \mae\big(\GenSrcTgt(\image, \dmapSrc), \dmapSrc\big), \\
\Reg^{\tgtdown}_{\text{range}}(\dmap^{\tgtdown}) &=  \mae\big(\GenTgtSrc(\image, \dmap^{\tgtdown}), \dmap^{\tgtdown}\big). \nonumber
\end{align}
We additionally aim to prevent the reverse translation network~$\GenSrcTgt$ from distorting depth images in the clean downsampled~$\tgtdown$ by introducing the \textit{identity regularization term}
\begin{equation}
\Reg_{\text{idt}}(\dmap^{\tgtdown})
 = \mae(\GenSrcTgt(\image, \dmap^{\tgtdown}), \dmap^{\tgtdown}).
\end{equation}

\begin{figure}[t]
     \begin{tikzpicture}
        \node[inner sep=0] (image) {\includegraphics[width=\linewidth]{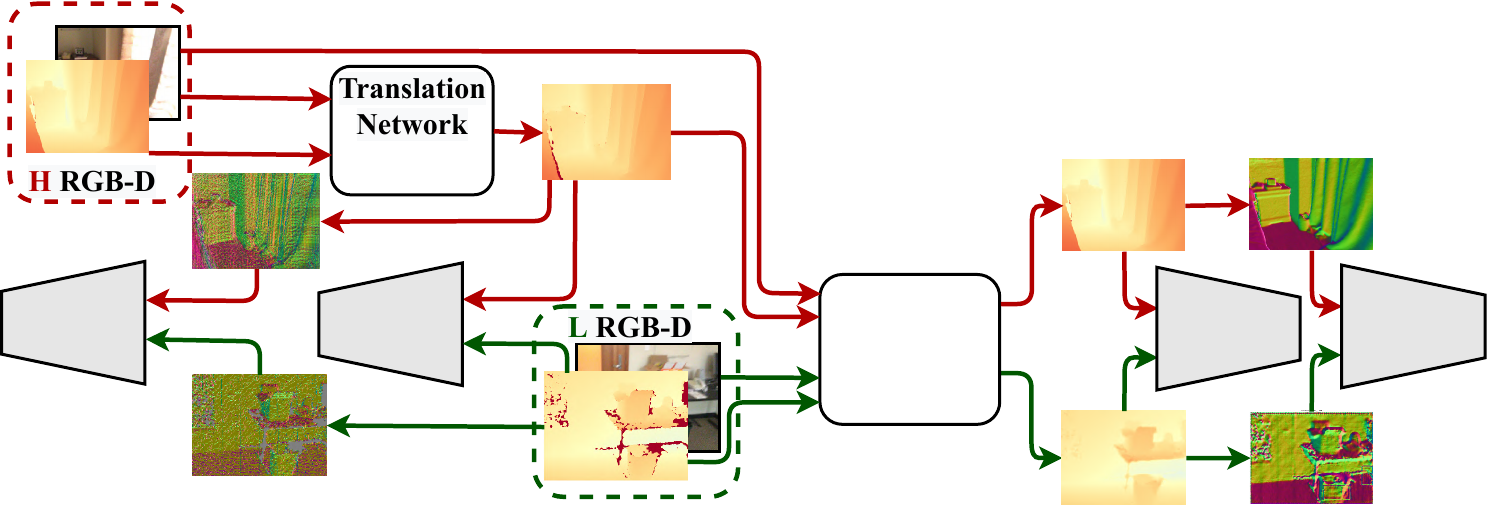}};
        \begin{scope}[shift=(image.north west), x={(image.north east)},y={(image.south west)}]
            \node at (0.95, .65) {\small{$\discNmapTgt$}};
            \node at (0.8253, .65) {\small{$\discDmapTgt$}};
            \node at (.287, .32) {\scriptsize{$\GenTgtSrc$}};
            \node at (.61, .67) {\scriptsize{$\GenSrcTgt$}};
            \node at (0.257, .65) {\small{$\discDmapSrc$}};
            \node at (0.05, .65) {\small{$\discNmapSrc$}};
        \end{scope}
    \end{tikzpicture}
    \caption{During training of the the translation network, we use normals-based discriminator networks in addition to depth-based discriminators; this leads to improved performance in our evaluation.}
    \label{fig:fig_translation_component_training}
\end{figure}
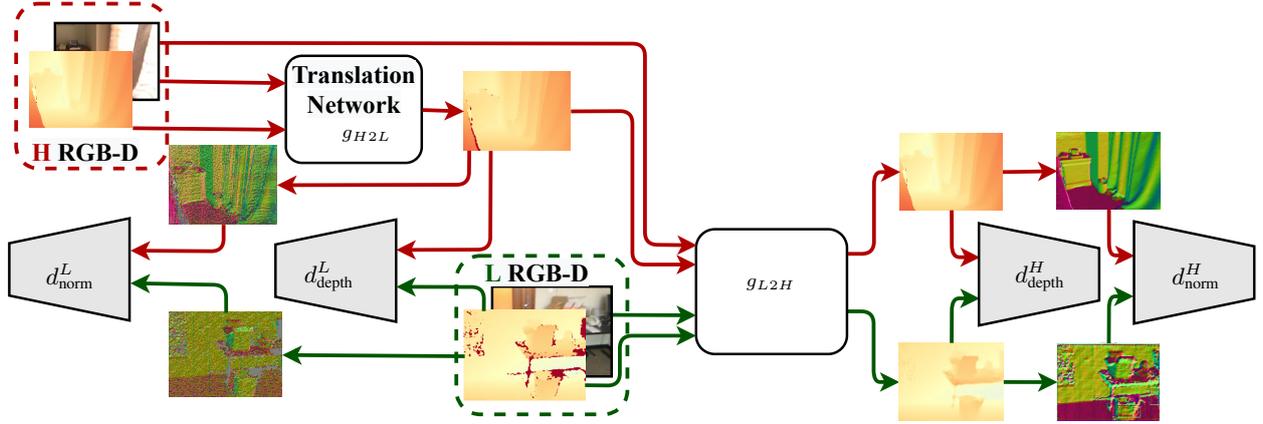

The system is trained by optimizing the following final translation objective:
\begin{align}
\label{eq:full_translation_objective}
\Loss_{\text{trans}}
 &= \Loss_{\text{adv}}
 + \lambda_{\text{cycle}} \Loss_{\text{cycle}} \\
 &+ \lambda^{\src}_{\text{range}} \Reg^{\src}_{\text{range}}
 + \lambda^{\tgtdown}_{\text{range}} \Reg^{\tgtdown}_{\text{range}}
 + \lambda_{\text{idt}} \Reg_{\text{idt}}. \nonumber
\end{align}
For brevity, we specify formulation for our adversarial loss $\Loss_{\text{adv}}$ in the Supplemental in Section~\ref{supp:sec:translation} as Equation~\eqref{eq:translation-adv-losses}.
We describe the architecture for our models and hyperparameter choices in the experimental Section~\ref{experiments:setup}.


\section{Methodology and~Data Construction for~Evaluation of~Paired and~Unpaired Approaches}
\label{sec:benchmark}

The goals of our evaluation are
(1) to compare our unpaired depth SR method with paired and unpaired methods on as equal terms as possible,
and (2) to study the performance of our method in a realistic unpaired training scenario.
For this, we develop a framework that enables directly performing such comparisons; we use this framework for creating a benchmark based on two datasets containing RGB-D scans of indoor scenes:
ScanNet~\cite{ScanNet:Dai2017}, which contains scans captured with Structure depth sensor,
and InteriorNet~\cite{InteriorNet:Li2018}, which contains high-quality simulated RGB-D scans.

\vspace{0.5em}
\noindent \emph{Dataset Construction Methodology.}
We start from a paired dataset containing aligned low- ($\src$) and high-quality RGB-D images (such as $\tgtdown$ or $\tgt$) of the same 3D environments (scenes) and adapt it as follows (see Figure~\ref{fig:fig_evaluation_datasets}):


\begin{figure}[h]
\centering
\includegraphics[width=0.6\columnwidth]{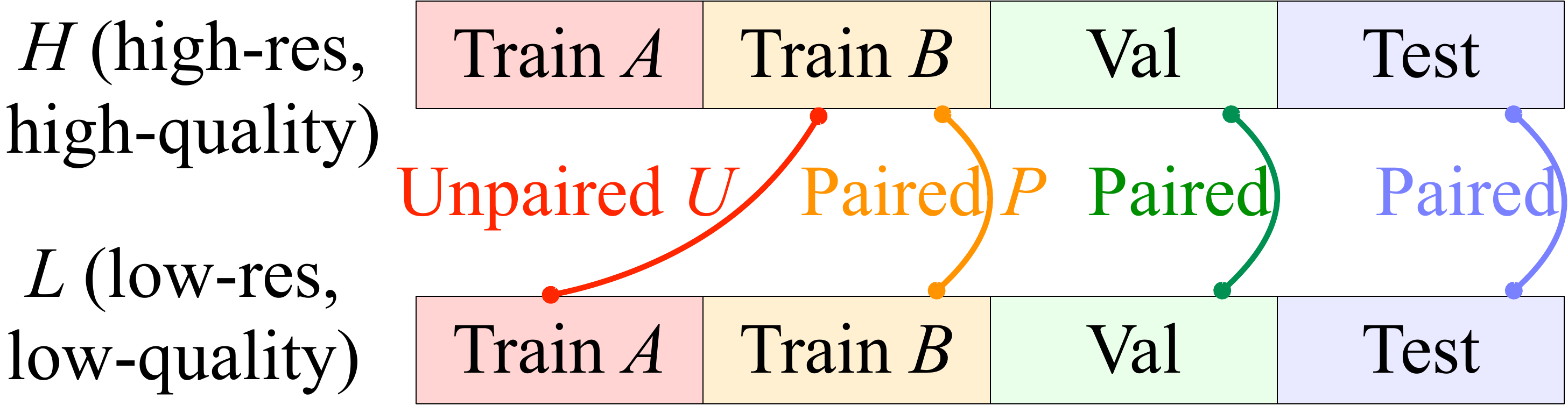}
\caption{Unpaired methods are trained on disjoint parts A and B, while paired ones have training pairs.}
\label{fig:fig_evaluation_datasets}
\end{figure}


\begin{enumerate}[topsep=0pt,itemsep=-1ex,partopsep=1ex,parsep=1ex]
    \item We split the dataset into training, validation and test sets, each using a different subset of scenes.
    \item We further split the training set into two disjoint parts, also using distinct scenes, Train~\emph{A} and Train~\emph{B}.
    \item The unpaired training set~$U$  includes low-quality images in Train~\emph{A} and high-quality images in Train~\emph{B}.
    \item The paired training set~$P$ combines low-quality images in Train~\emph{B} and high-quality images in Train~\emph{B}.
    \item Test set $T$, obtained at step 1, is the same for both paired and unpaired methods.
\end{enumerate}
This approach enables directly comparing unsupervised methods trained on~$U$ with supervised methods trained on $P$.
We further develop a depth SR and enhancement benchmark, that we describe below.

\vspace{0.5em}
\noindent \emph{Benchmark Varieties.}
Our benchmark consists of three parts that serve distinct purposes, as summarized by Table~\ref{tab:datasets_traits}.
\textit{ScanNet-RenderScanNet} is built from raw sensor RGB-D images from ScanNet and high-quality depth rendered from 3D reconstructions of ScanNet scenes.
This benchmark aims to provide uniform training data for both paired and unpaired methods, and is split into three sub-parts \textit{Train~A,} \textit{Train~B,} and \textit{Val,} following the data construction framework.
Additionally, training on \textit{ScanNet-RenderScanNet} allows to study unpaired methods in a controlled training scenario when the low- and high-quality data differ only in quality but not in semantics or other properties which could impair training stability.
\textit{ScanNet-InteriorNet} contains real-world sensor data from ScanNet and synthetic high-quality data from InteriorNet.
With this benchmark, we aim to evaluate unpaired methods in a realistic training scenario when the low- and high-quality data differ not only in quality, but also in semantic content, distribution of depth values, etc.
\textit{Testing dataset} is a hold-out set for evaluation of both paired and unpaired depth map super-resolution methods; it consists of
 full-sized sensor RGB-D images from ScanNet and their respective high-quality renders.

\begingroup

\renewcommand{\arraystretch}{1.15}
\setlength\tabcolsep{\widthof{0}*\real{.99}}

\begin{table}[t]
\centering
\begin{tabular}{
l
l
c
c
c
r}
\toprule
\textbf{Split}  &
    \textbf{Set}    &
    \textbf{Source}         & 
    \textbf{RGB}            &   
    \textbf{Depth}          &  
    \textbf{Volume}         \\ 
\midrule
\multicolumn{6}{l}{\textit{ScanNet-RenderScanNet} (paired and unpaired)} \\
\textit{Train}\,$A$                                 & 
    $\src$                                          &
    ScanNet~\cite{ScanNet:Dai2017}                  & 
    $1280 \times 960$                               & 
    $640 \times 480$                                & 
    6\,221                                          \\
\textit{Train}\,$B$                                 & 
    $\src$                                          &
    ScanNet~\cite{ScanNet:Dai2017}                  & 
    $1280 \times 960$                               & 
    $640 \times 480$                                & 
    6\,221                                          \\
                                                    & 
    $\tgt$                                          &
    Rendered~\cite{BundleFusion:Dai2017}            & 
    $1280 \times 960$                               & 
    $1280 \times 960$                               & 
    6\,221                                          \\
\textit{Val}                                        & 
    $\src$                                          &
    ScanNet~\cite{ScanNet:Dai2017}                  & 
    $1280 \times 960$                               & 
    $640 \times 480$                                & 
    2\,945                                          \\
                                                    & 
$\tgt$                                              &
    Rendered~\cite{BundleFusion:Dai2017}            & 
    $1280 \times 960$                               & 
    $1280 \times 960$                               & 
    2\,945                                          \\
\midrule
\multicolumn{6}{l}{\textit{ScanNet-InteriorNet} (unpaired only)}    \\
\textit{Train}\,$A$                                 & 
    $\src$                                          &
    \multicolumn{4}{l}{\textit{Train}\,$A$ ($\src$) in 
    \textit{ScanNet-RenderScanNet}}                 \\
\textit{Train}\,$B$                                 & 
    $\tgt$                                          &
    InteriorNet~\cite{InteriorNet:Li2018}           & 
    $1280 \times 960$                               & 
    $1280 \times 960$                               & 
    13\,744                                         \\
\midrule
\multicolumn{6}{l}{\textit{Testing dataset}}        \\
\textit{Test}                                       & 
    $\src$                                          &
    ScanNet~\cite{ScanNet:Dai2017}                  & 
    $1280 \times 960$                               & 
    $640 \times 480$                                & 
    501 \\
                                                    & 
    $\tgt$                                          &
    Rendered~\cite{BundleFusion:Dai2017}            & 
    $1280 \times 960$                               & 
    $1280 \times 960$                               & 
    501                                             \\
\bottomrule
\end{tabular}%
\caption{Summary of the used training and validation datasets.
\enquote{Rendered~\cite{BundleFusion:Dai2017}} refers to rendering depth images from 3D reconstructions of ScanNet scenes obtained using BundleFusion~\cite{BundleFusion:Dai2017}.}
\label{tab:datasets_traits}
\end{table}

\endgroup

\section{Experimental Evaluation}
\label{sec:experiments}

\subsection{Experimental Setup}
\label{experiments:setup}

We evaluate our method in depth map super-resolution with a scaling factor $k = 2$,
training it on either \textit{ScanNet-RenderScanNet} or \textit{ScanNet-InteriorNet,} and testing on \textit{Testing dataset.}
As our method is capable of performing depth enhancement at the original spatial resolution, we present an additional evaluation on depth enhancement task with~$k = 1$.
Finally, we study the individual contributions of different components of our method in an ablation study.
We specify the experimental setup here and give details specific to each experiment in respective Sections~\ref{experiments:sr-scanner-renderscannet}--\ref{experiments:ablation}.

\vspace{0.5em}
\noindent \textit{Network Architectures.}
Translation networks~$\GenTgtSrc$ and~$\GenSrcTgt$ accept an RGB and a depth image as input and produce a depth image as output. 
Both networks share the same architecture that consist of convolutional RGB and depth feature encoders computing features that are concatenated and processed by 9 ResNet~\cite{ResNet:He2015} blocks, and a convolutional decoder that produces a final depth image. 
For feature normalization in these models, we apply GroupNorm~\cite{GroupNorm:Wu2020}.
All discriminators follow the CycleGAN~\cite{CycleGAN:Zhu2017} discriminator architecture.
We use Spectral Normalization~\cite{SpectralNorm:Miyato2018} for weight regularization but perform no feature normalization in discriminators.

The RGB Guidance Network~$\fimg$ architecturally includes a feature extractor (2 downsampling blocks, 6 ResNet blocks, and 2 upsampling blocks) and the depth prediction (a vanilla U-Net~\cite{UNet:Ronneberger2015}) sub-networks.
We use Instance Normalization~\cite{DBLP:journals/corr/UlyanovVL16} in both these parts.

The enhancement network~$\fenh$ architecturally is a vanilla U-Net.

\vspace{0.5em}
\noindent \textit{Data Preprocessing and Augmentation.}
All super-resolution methods in our experiments are based on fully-convolutional neural networks and can be naturally trained on crops of arbitrary size, and we build \textit{Train~B} and \textit{Val} sub-parts from crops instead of full-sized images in order to increase the size of these subsets, as we explain further.

For the three datasets (ScanNet, RenderScanNet, and InteriorNet), we select RGB-D frames with the maximum depth value not exceeding 5.1\,m, corresponding to the 85th depth range percentile in ScanNet~\cite{ScanNet:Dai2017}. 
Such filtering is consistent with the depth range where commercial depth sensors have sufficient precision while simultaneously allowing to reduce discrepancy better between images in synthetic and real datasets and providing common threshold for data normalization, which is a common practice for deep learning.

The depth fusion process \cite{BundleFusion:Dai2017} is not perfect, and rendering of reconstructed meshes has been shown to produce local misalignments~\cite{RecScanNet:Jeon2018} and create unnatural regions in output images. 
To reduce this effect, we applied a filtering procedure similar to the one used in~\cite{RecScanNet:Jeon2018}: we extracted 320$\times$320 and 640$\times$640 patches from input RGB-D frames and measured structured similarity~\cite{SSIM:Zhou2004} (SSIM) between the raw and rendered depth patch and its downsampled version, selecting images with SSIM greater than 0.8, and discarding pairs if the rendering produces misaligned result relatively to the sensor depth. We also store full frames with at least one patch selected.

To improve hole inpainting performance, we randomly add $N$ rectangular holes to each training RGB-D frame, sampling $N$ uniformly in [10, 75] range. 
The holes have random sizes $(h_n, w_n)$ where $h_n$ is uniformly sampled in the range ${\tgt}/128 \ldots {\tgt}/8$, and $w_n$ from $W/128 \ldots W/8$  where $({\tgt}, W)$ represent the dimensions of the input depth image. 
We perform this during each training iteration with probability 0.9.

\begingroup

\renewcommand{\arraystretch}{1.15}
\setlength\tabcolsep{\widthof{0}*\real{.3}}

\def\u{\underline}
\def\b{\textbf}

\begin{table}
\centerline{
\begin{threeparttable}
\begin{tabular}{l r r r r r r r }
\toprule
Method                  & 
    \rmse$\downarrow$   & 
    \rmse[h]            & 
    \rmse[d]            & 
    \mae                & 
    \mae[h]             & 
    \mae[d]             & 
    \mse[v]             \\ 
\midrule
SRFBN~\cite{SRFBN:Li_2019}\tnote{\textasteriskcentered}                         & 363.5             & 1435.5            & 129.0            & 109.3            & 1392.1            & \b{27.1}         & 25.8         \\
MS-PFL~\cite{MS-PFL:Chuhua2020}\tnote{\textasteriskcentered}                    & 299.5             & 1163.8            & 117.7            & 93.7             & 1110.6            & \u{29.3}         & \u{24.9}     \\
Gu~\cite{Gu2020_TIP} $+$ SRBFN~\cite{SRFBN:Li_2019}\tnote{\textasteriskcentered}& \u{107.7}         & 194.1             & 94.7             & 56.8             & \u{121.2}         & 51.7             & 28.1         \\
SRFBN~\cite{SRFBN:Li_2019}\tnote{\textsection}                                  & 76.9              & 176.4             & 58.5             & 22.2             & 92.5              & 16.4             & 16.4         \\ 
MS-PFL~\cite{MS-PFL:Chuhua2020}\tnote{\textsection}                             & 75.5              & 168.4             & 60.4             & 30.2             & 113.3             & 23.8             & 16.1         \\
\midrule
Bicubic $+$ Gu~\cite{Gu2020_TIP}                                                & 114.7             & 323.0             & \u{78.5}         & \u{56.3}         & 246.5             & 43.1             & 52.0         \\
Gu~\cite{Gu2020_TIP} $+$ Bicubic                                                & 108.0             & \u{193.8}         & 95.0             & 57.1             & 121.4             & 52.1             & 28.5         \\
\midrule
UDSR (Ours)                                                                     & \b{86.2}          & \b{172.6}         & \b{74.7}         & \b{45.5}         & \b{113.8}         & 40.6             & \b{20.2}     \\ 
\bottomrule
\end{tabular}
\begin{tablenotes}
\footnotesize
    \item[\textsection] supervised algorithms trained on ($\src$, $\tgt$)
    \item[\textasteriskcentered] supervised algorithms trained on ($\tgtdown$, $\tgt$) 
    \item non-marked methods Gu\,\textit{et\,al.}~\cite{Gu2020_TIP} and UDSR are unpaired
\end{tablenotes}
\end{threeparttable}
}
\caption{Our statistical evaluation of depth SR on \textit{ScanNet-RenderScanNet} benchmark demonstrates that our UDSR outperforms all unpaired methods and quantitatively approaches the supervised SRFBN~\cite{SRFBN:Li_2019} and MS-PFL~\cite{MS-PFL:Chuhua2020}.
Lower = better; we show $\mse[v] \times 10^{-2}$.}
\label{tab:sr_scannet_renderscannet}
\end{table}

\endgroup

\vspace{0.5em}
\noindent \textit{Quality Measures.}
We calculate seven performance measures by comparing the super-resolved or enhanced depth images against their high-quality counterparts.
\rmse, the root mean squared error, emphasizes large deviations.
\mae, the mean absolute error, quantifies an average error without taking outliers into account.
\rmse[h] and \mae[h], the errors averaged over pixels with missing input depth value, assess the inpainting performance of the method.
\rmse[d] and \mae[d], the errors averaged over pixels with defined input depth value, measure the quality of the method in regions with valid input,  and are well-suited for the evaluation of approaches that do not perform inpainting.
Finally, \mse[v] defined in~\cref{eqn:actual_mse_v} measures the perceptual similarity between the 3D surface represented using an output depth map and the 3D surface represented by a reference depth map.
In all calculations, we exclude pixels with unknown reference depth values.
We report all metrics except \mse[v] in millimeters.

\vspace{0.5em}
\noindent \textit{Training Competitor Methods.}
For a fair comparison between competitors and our approach we re-train all competitor methods on the datasets available in our benchmark; we do not use datasets available from respective authors as they are unable to provide equal conditions for our comparisons.


\begingroup

\renewcommand{\arraystretch}{1.15}
\setlength\tabcolsep{\widthof{0}*\real{.3}}

\def\u{\underline}
\def\b{\textbf}

\begin{table}
\centerline{
\begin{threeparttable}
\begin{tabular}{l r r r r r r r }
\toprule
Method                  & 
    \rmse$\downarrow$   & 
    \rmse[h]            & 
    \rmse[d]            & 
    \mae                & 
    \mae[h]             & 
    \mae[d]             & 
    \mse[v]             \\ 
\midrule
Bicubic $+$ Gu~\cite{Gu2020_TIP}                    & 241.2        & 893.1        & 109.0        & 93.5         & 771.5        & 50.5         & 47.9        \\ 
Gu~\cite{Gu2020_TIP} $+$ SRFBN~\cite{SRFBN:Li_2019} & 107.2        & 298.2        & \u{67.8}     & \u{45.7}     & 210.0        & \u{33.1}     & 42.5        \\
Gu~\cite{Gu2020_TIP} $+$ Bicubic                    & \u{107.1}    & \u{297.7}    & 67.9         & 46.1         & \u{209.9}    & 33.6         & \u{36.7}    \\
\midrule
UDSR (Ours)                                         & \b{81.1}     & \b{197.9}    & \b{61.0}     & \b{28.0}     & \b{123.5}    & \b{20.6}     & \b{26.1}    \\
\bottomrule
\end{tabular}
\end{threeparttable}
}
\caption{For an unpaired scenario \textit{ScanNet-InteriorNet} with geometric and semantic differences in source and target datasets, our method outperforms state-of-the-art depth enhancement~\cite{Gu2020_TIP} coupled with trained~\cite{SRFBN:Li_2019} and untrained depth upsampling steps, across all quality measures we computed.
Lower = better; we show $\mse[v] \times 10^{-2}$.}
\label{tab:sr_scannet_interiornet}
\end{table}

\endgroup


\subsection{Super-Resolution on ScanNet-RenderScanNet.}
\label{experiments:sr-scanner-renderscannet}

\noindent \textit{Methods and Training.}
We compare our method with three other learning-based methods.
MS-PFL~\cite{MS-PFL:Chuhua2020} is a state-of-the-art supervised depth SR method based on progressive multi-scale fusion of features extracted from input depth and RGB images. 
SRFBN~\cite{SRFBN:Li_2019} is an established supervised method for RGB image super-resolution based on recurrent connections which allow to incorporate high-level information flow.
SRFBN often serves as a strong baseline for evaluating depth super-resolution methods (for instance, in~\cite{CA-IRL_SR:Song_2020,MS-PFL:Chuhua2020}).
Following~\cite{MS-PFL:Chuhua2020}, we modified SRFBN to take a single-channel depth tensor as input and produce a single-channel output instead of a three-channel RGB one.
Lastly, Gu~\etal~\cite{Gu2020_TIP} is the only existing unpaired method for depth enhancement with a state-of-the-art performance. 
We adapt this method for super-resolution as explained below.
%

To more fully characterize performance of supervised methods, we include results obtained by training these methods to predict clean high-resolution targets in~$\tgt$ using two distinct variants of input data (see \cref{tab:sr_scannet_renderscannet}).
In the first variant, we use synthesized low-resolution inputs from the downsampled~$\tgtdown$ as done commonly by depth SR literature; 
the second variant consists in using registered real-world inputs from the source~$\src$ available in our benchmarks.

To adapt the enhancement method of Gu~\etal for super-resolution, we combined it with upsampling in three ways.
For the first combination (we denote it \textquote{Bicubic + Gu}), we trained the method of Gu~\etal on bicubically upsampled sensor depth from the source~$\src$ as inputs
and high-resolution high-quality depth images from~$\tgt$ as targets,
and for testing applied the method to bicubically upsampled sensor depth.
For the second (\textquote{Gu + Bicubic}), we trained the method on sensor depth maps from~$\src$ as inputs
and downsampled high-quality depth maps from the downsampled~$\tgtdown$ as targets,
and for testing applied bicubic interpolation to the output of the method.
For the last combination (\textquote{Gu + SRFBN}), we again trained the method of Gu~\etal on depth maps from~$\src$ and~$\tgtdown$,
but for upsampling during testing we used SRFBN trained on pairs from~$\tgtdown$ and~$\tgt$.

\vspace{0.5em}
\noindent \textit{Results.}
We display statistical evaluation results in~\cref{tab:sr_scannet_renderscannet} and visualize example predictions in~\cref{fig:sr_render}.
Our method outperforms all variants of the unpaired method of Gu~\etal adapted for depth SR both quantitatively and qualitatively.
All variants, particularly \textquote{Bicubic + Gu} where enhancement follows upsampling, produce surfaces with step-like artefacts, which is illustrated by the greyish color of normal maps, and which leads to a low perceptual quality (\mse[v]) of the result.
Variants where upsampling follows enhancement, \ie, \textquote{Gu + Bicubic} and \textquote{Gu + SRFBN}, produce very similar results, so we only show the qualitative results for the latter.

Compared to paired methods trained in the commonly used scenario on downsampled inputs from~$\tgtdown$, our method achieves lower \rmse[d] and \mse[v] but a higher \mae[d].
As illustrated by visuals of the normal maps, compared to our algorithm, these methods produce significantly noisier surface, almost as noisy as the input sensor data; this is likely due to the domain shift between the synthetic input during training and the real-world sensor input during testing.
This aligns with our initial motivation of exploring unpaired methods against the lack of representative paired data reflecting real-world input.

Compared to paired methods trained in the second scenario on data from~$\src$ and~$\tgt$, our unpaired method performs only slightly worse quantitatively.
At the same time, qualitative results of these methods contain artefacts that are not present in the predictions of our method; for instance,  SRFBN produces ringing artefacts around object boundaries, and MS-PFL tends to output an over-smoothed depth.

\begingroup

\renewcommand{\arraystretch}{1.15}
\setlength\tabcolsep{\widthof{0}*\real{.3}}

\def\u{\underline}
\def\b{\textbf}

\begin{table}
\centerline{
\begin{tabular}{l r r r r r r r}
\toprule
Method                  & 
    \rmse$\downarrow$   & 
    \rmse[h]            & 
    \rmse[d]            & 
    \mae                & 
    \mae[h]             & 
    \mae[d]             & 
    \mse[v]             \\ 
\midrule
LapDEN~\cite{RecScanNet:Jeon2018}       & 76.0         & 165.6        & 58.4         & 22.6         & 92.7         & 15.9         & 7.7 \\
\midrule
CycleGAN~\cite{CycleGAN:Zhu2017}        & 416.6        & 450.6        & 407.5        & 391.2        & 389.4        & 392.5        & \u{13.3} \\ 
U-GAT-IT~\cite{U-GAT-IT:Kim2020}        & 292.8        & 508.0        & 261.3        & 249.5        & 434.5        & 235.6        & 14.2 \\
NiceGAN~\cite{NiceGAN:Chen2020}         & 156.5        & 410.2        & 111.5        & 97.6         & 324.8        & 79.5         & 19.5 \\
Gu\,\textit{et\,al.}~\cite{Gu2020_TIP}  & \u{108.3}    & \u{193.9}    & \u{94.3}     & \u{57.1}     & \u{120.2}    & \u{51.7}     & 33.1 \\
\midrule 
UDSR (Ours)                             & \b{77.1}     & \b{169.6}    & \b{63.3}     & \b{33.0}     & \b{110.1}    & \b{27.3}     & \b{11.9} \\
\bottomrule
\end{tabular}
}
\caption{Enhancement performance statistics using the \textit{ScanNet-RenderScanNet} benchmark indicates that our UDSR significantly outperforms all existing unpaired methods and quantitatively approaches the supervised LapDEN~\cite{RecScanNet:Jeon2018}.
Lower = better; we show $\mse[v] \times 10^{-2}$. }
\label{tab:enhancement_scannet_renderscannet}
\end{table}

\endgroup

\subsection{Super-Resolution on ScanNet-InteriorNet.}
\label{experiments:sr-scanner-interiornet}

\noindent \textit{Methods and Training.}
For evaluation using the \textit{ScanNet-InteriorNet} benchmark, we compared our method with the three adaptations of the method of Gu~\etal described above.

\vspace{0.5em}
\noindent \textit{Results.}
Evaluation results are shown quantitatively in~\cref{tab:sr_scannet_interiornet} and visually in~\cref{fig:sr_interior}.
Our method outperforms all three variants of Gu~\etal which suffer from the same issues as during training on \textit{ScanNet-RenderScanNet}.
Notably, using ideal target data from InteriorNet leads to a similar or improved quantitative performance for our method, compared to using renders of ScanNet reconstructions.


\subsection{Enhancement on ScanNet-RenderScanNet.}
\label{experiments:enh-scanner-interiornet}

\noindent \textit{Methods and Training.}
In the task of depth enhancement with no change in spatial resolution we compared our method with five other learning-based methods: 
the unpaired method of Gu~\etal,
LapDEN~\cite{RecScanNet:Jeon2018}, a state-of-the-art supervised method for depth enhancement based on deep Laplacian pyramid network, 
and three unpaired methods for RGB image-to-image translation, 
CycleGAN~\cite{CycleGAN:Zhu2017}, U-GAT-IT~\cite{U-GAT-IT:Kim2020}, and NiceGAN~\cite{NiceGAN:Chen2020}.
We modified them to take a single-channel depth tensor as input and produce a single-channel output,
and for CycleGAN we additionally trained a version with four-channel RGB-D input.

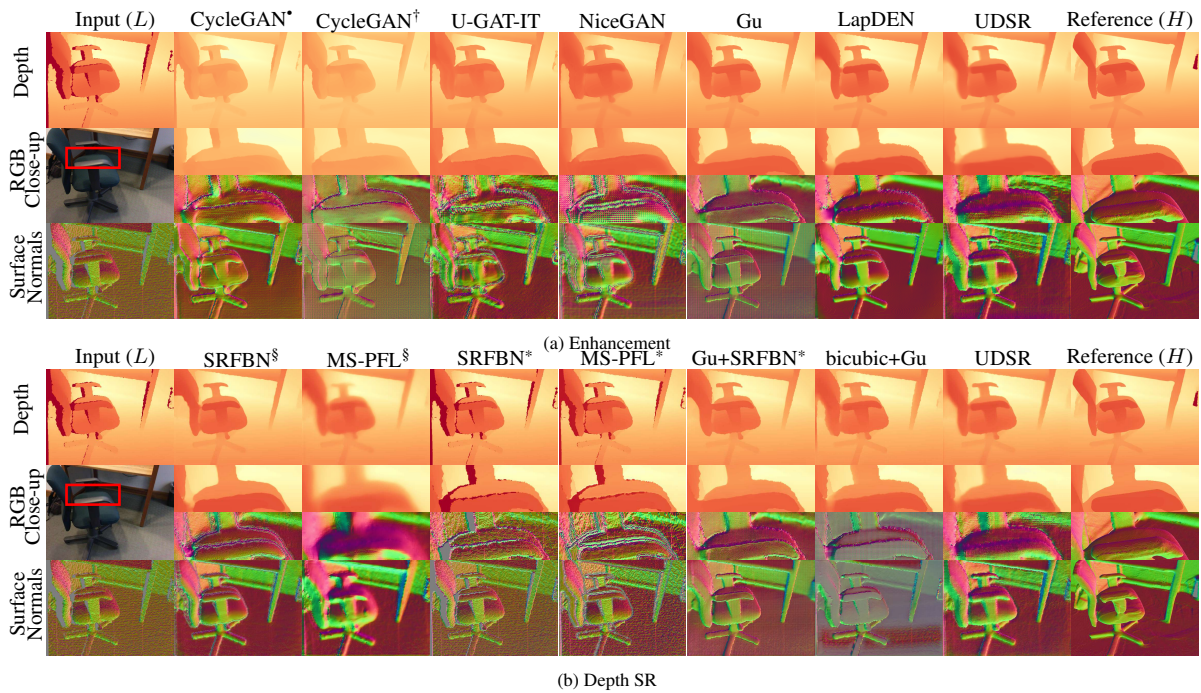
\begin{figure*}
    \begin{subfigure}{.97\textwidth}
        \begin{tikzpicture}
            \node[inner sep=0] (image) {\includegraphics[width=.95\linewidth]{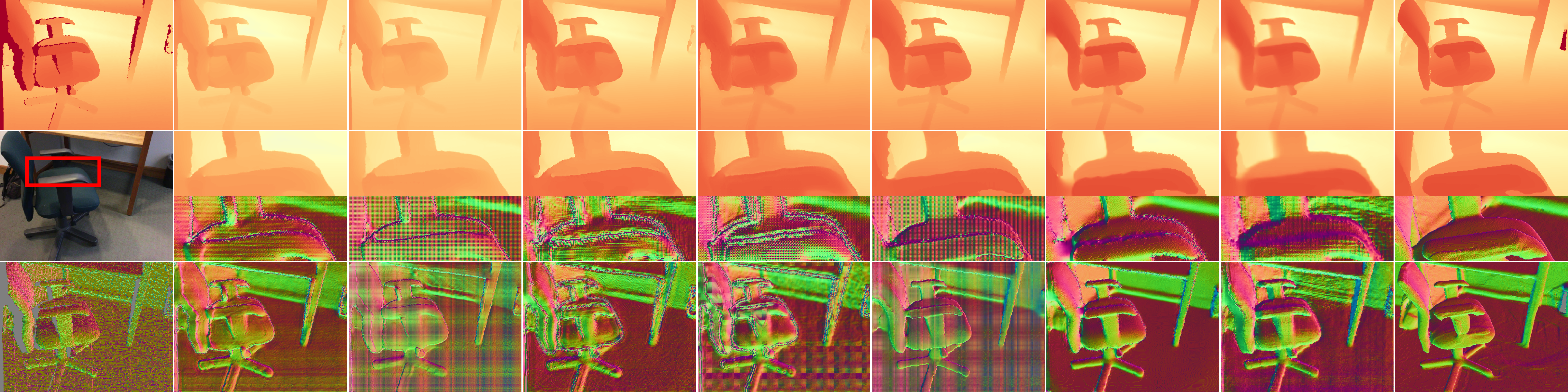}};
            \begin{scope}[shift=(image.north west), x={(image.north east)},y={(image.south west)}]
                \node[rotate=90] at (-.02, .14) {\footnotesize {Depth}};
                \node[rotate=90] at (-.025, .48) {\footnotesize {RGB}};
                \node[rotate=90] at (-.01, .48) {\footnotesize {Close-up}};
                \node[rotate=90] at (-.025, .82) {\footnotesize {Surface}};
                \node[rotate=90] at (-.01, .82) {\footnotesize {Normals}};
                \node at (.06, -.04) {\footnotesize {Input ($\src$)}};
                \node at (.17, -.04) {\footnotesize {$\text{CycleGAN}^{\text{\textbullet}}$}};
                \node at (.28, -.04) {\footnotesize {$\text{CycleGAN}^{\text{\textdagger}}$}};
                \node at (.39, -.04) {\footnotesize {$\text{U-GAT-IT}$}};
                \node at (.50, -.04) {\footnotesize {$\text{NiceGAN}$}};
                \node at (.61, -.04) {\footnotesize {Gu}};
                \node at (.72, -.04) {\footnotesize {LapDEN}};
                \node at (.83, -.04) {\footnotesize {UDSR}};
                \node at (.94, -.04) {\footnotesize {Reference ($\tgt$)}};
            \end{scope}
    	\end{tikzpicture}
    	\caption{Enhancement}
    	\label{fig:enh_render}
	\end{subfigure}
    \begin{subfigure}{.97\textwidth}
    	\begin{tikzpicture}
            \node[inner sep=0] (image) {\includegraphics[width=.95\linewidth]{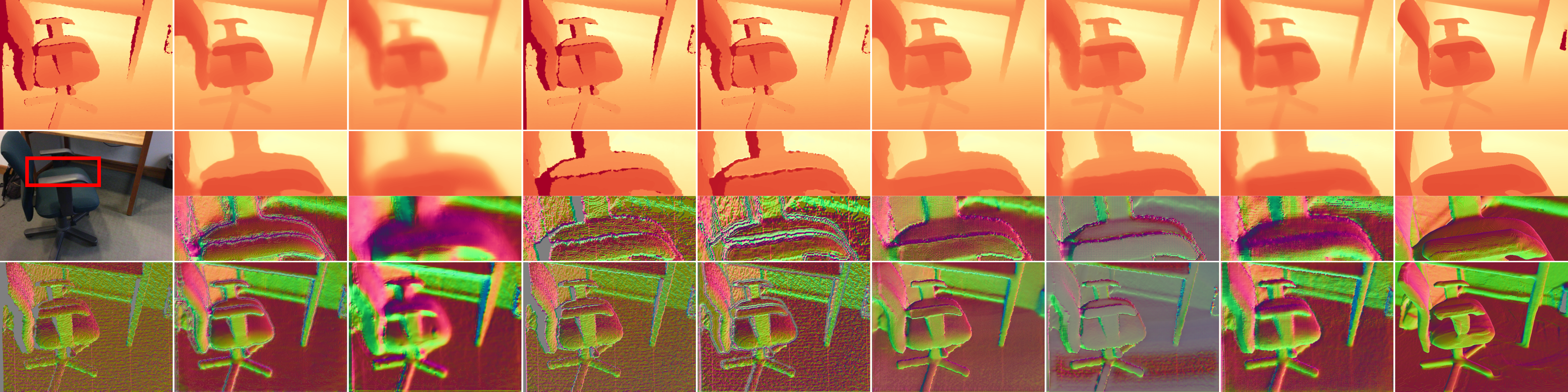}};
            \begin{scope}[shift=(image.north west), x={(image.north east)},y={(image.south west)}]
                \node at (.06, -.04) {\footnotesize {Input ($\src$)}};
                \node at (.17, -.04) {\footnotesize {$\text{SRFBN}^\text{\textsection}$}};
                \node at (.28, -.04) {\footnotesize {$\text{MS-PFL}^{\text{\textsection}}$}};
                \node at (.39, -.04) {\footnotesize {$\text{SRFBN}^\text{\textasteriskcentered}$}};
                \node at (.50, -.04) {\footnotesize {$\text{MS-PFL}^{\text{\textasteriskcentered}}$}};
                \node at (.61, -.04) {\footnotesize {Gu+$\text{SRFBN}^\text{\textasteriskcentered}$}};
                \node at (.72, -.04) {\footnotesize {bicubic+Gu}};
                \node at (.83, -.04) {\footnotesize {UDSR}};
                \node at (.94, -.04) {\footnotesize {Reference ($\tgt$)}};
                \node[rotate=90] at (-.02, .14) {\footnotesize {Depth}};
                \node[rotate=90] at (-.025, .48) {\footnotesize {RGB}};
                \node[rotate=90] at (-.01, .48) {\footnotesize {Close-up}};
                \node[rotate=90] at (-.025, .82) {\footnotesize {Surface}};
                \node[rotate=90] at (-.01, .82) {\footnotesize {Normals}};
            \end{scope}
    	\end{tikzpicture}
    	\caption{Depth SR}
    	\label{fig:sr_render}
	\end{subfigure}
    \caption{Enhancement and depth SR on \textit{ScanNet-RenderScanNet} benchmark.
    \textsuperscript{\textsection} and \textsuperscript{\textasteriskcentered} mark the models trained on ($\src$, $\tgt$) and ($\tgtdown$, $\tgt$) pairs respectively.
    \textsuperscript{\textbullet} and \textsuperscript{\textdagger} mark the models with the depth map input and depth map concatenated with RGB respectively.You can find additional results supplementary materials in ~\cref{fig:sup_enh_scannet_rscannet} and ~\cref{fig:sup_sr_scannet_rscannet}}
\end{figure*}

\vspace{0.5em}
\noindent \textit{Results.}
The evaluation results are shown in~\cref{tab:enhancement_scannet_renderscannet} and in~\cref{fig:enh_render}.
Quantitatively, our method outperforms the other unpaired methods in all measures.
Notably, even for a harder task of $2 \times$ depth SR, our method achieves higher scores in all measures except \mse[v] compared to other unpaired methods on the task of depth enhancement (no change in spatial resolution).
Qualitatively, our method successfully performs denoising, hole inpainting, and yields surface normals close to the target while the other unpaired methods suffer from various artefacts.
Gu~\etal produces the result with step-like artefacts, which is indicated by the greyish color of the normal map and a high value of~\mse[v];
NiceGAN and U-GAT-IT suffer from ringing artefacts around the object boundaries;
U-GAT-IT and CycleGAN fail to preserve the correct absolute depth value, which is indicated by the shifted colors in depth image visualization and high values of \rmse- and \mae-based measures.
Compared to the paired LapDEN, our method performs slightly worse quantitatively
and produces the result with slightly noisier surface.

\begingroup

\renewcommand{\arraystretch}{1.15}
\setlength\tabcolsep{\widthof{0}*\real{.3}}

\def\u{\underline}
\def\b{\textbf}

\begin{table}
\centerline{
\begin{threeparttable}
\begin{tabular}{l r r r r r r r }
\toprule
Method                  & 
    \rmse$\downarrow$   & 
    \rmse[h]            & 
    \rmse[d]            & 
    \mae                & 
    \mae[h]             & 
    \mae[d]             & 
    \mse[v]             \\ 
\midrule
NiceGAN~\cite{NiceGAN:Chen2020}\tnote{\textbullet}  & 2063.7       & 2824.6       & 1984.3      & 1860.0    & 2759.8    & 1787.0    & 59.5      \\
U-GAT-IT~\cite{U-GAT-IT:Kim2020}\tnote{\textbullet} & 1300.2       & 1064.4       & 1311.4      & 1254.7    & 999.3     & 1276.4    & \u{24.4}  \\
CycleGAN~\cite{CycleGAN:Zhu2017}\tnote{\textbullet} & 405.9        & 640.0        & 359.5       & 356.8     & 566.6     & 336.6     & 48.2      \\
CycleGAN~\cite{CycleGAN:Zhu2017}\tnote{\textdagger} & 471.9        & 523.0        & 457.8       & 425.1     & 457.7     & 421.9     & 43.9      \\
Gu \etal~\cite{Gu2020_TIP}                          & \u{107.3}    & \u{290.1}    & \u{66.8}    & \u{46.1}  & \u{201.0} & \u{33.1}  & 44.4      \\ 
\midrule
UDSR (Ours)                                         & \b{80.8}     & \b{199.9}    & \b{59.1}    & \b{29.9}  & \b{124.8} & \b{22.2}  & \b{14.7}  \\
\bottomrule
\end{tabular}
\begin{tablenotes}
\footnotesize
    \item[\textbullet] input: depth image
    \item[\textdagger] input: depth image concatenated with RGB
\end{tablenotes}
\end{threeparttable}
}
\caption{Enhancement performance statistics for \textit{ScanNet-InteriorNet} scenario. 
Our method outperforms all competitors across all measures we computed.
Lower = better; we show $\mse[v] \times 10^{-2}$.}
\label{tab:enh_scannet_interiornet}
\end{table}

\endgroup

\subsection{Enhancement on ScanNet-InteriorNet.}
\label{experiments:enh-scanner-interiornet}

\noindent \textit{Methods and Training.}
We additionally perform evaluation of depth enhancement using the data available in the \textit{ScanNet-InteriorNet} benchmark, using the same methods as described previously.
We consider two input possibilities where we supply depth image or depth image concatenated with RGB as input to CycleGAN~\cite{CycleGAN:Zhu2017}; U-GAT-IT~\cite{U-GAT-IT:Kim2020}, and NiceGAN~\cite{NiceGAN:Chen2020} are trained using depth images only.

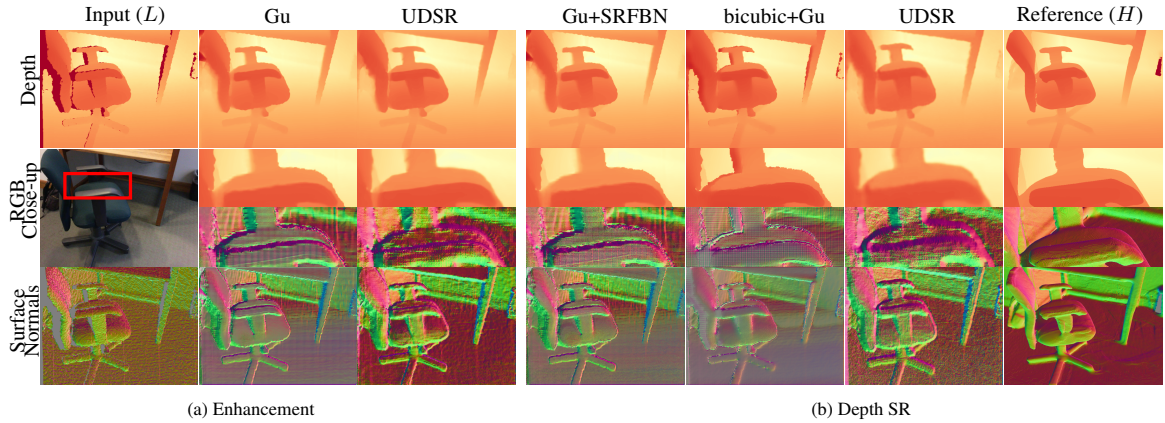
\begin{figure*}[t]
    \begin{subfigure}{.4\linewidth}
        \begin{tikzpicture}
            \node[inner sep=0] (image) {\includegraphics[width=.95\linewidth]{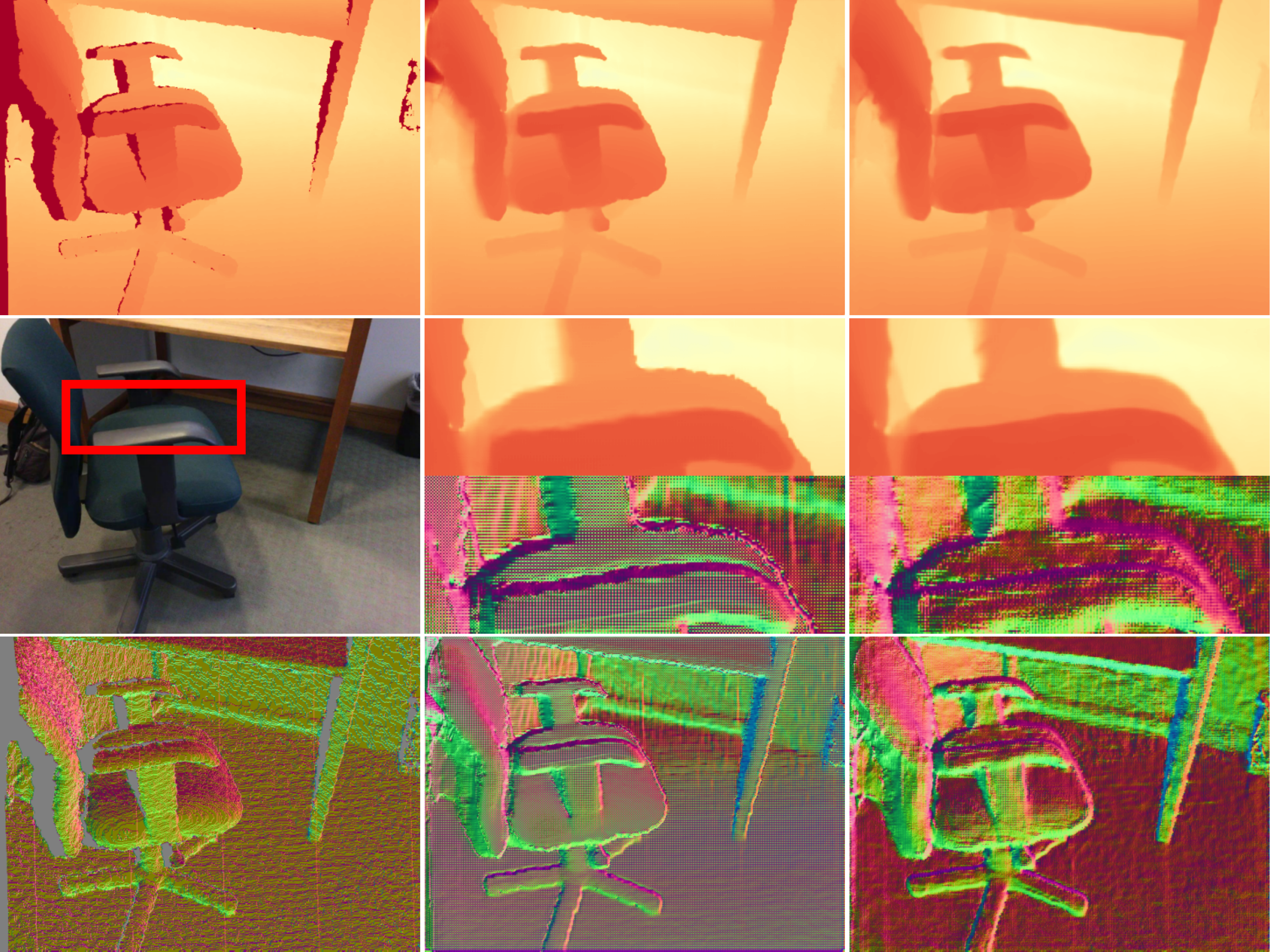}};
            \begin{scope}[shift=(image.north west), x={(image.north east)},y={(image.south west)}]
            \node[rotate=90] at (-.02, .14) {\footnotesize {Depth}};
            \node[rotate=90] at (-.047, .48) {\footnotesize {RGB}};
            \node[rotate=90] at (-.017, .48) {\footnotesize {Close-up}};
            \node[rotate=90] at (-.047, .82) {\footnotesize {Surface}};
            \node[rotate=90] at (-.017, .82) {\footnotesize {Normals}};
            \node at (.18, -.04) {\footnotesize {Input ($\src$)}};
            \node at (.5, -.04) {\footnotesize {Gu}};
            \node at (.82, -.04) {\footnotesize {UDSR}};
            \end{scope}
        \end{tikzpicture}
    	\caption{Enhancement}
    	\label{fig:enh_interior}
	\end{subfigure}
	\hspace{0.01\linewidth}
    \begin{subfigure}{.535\linewidth}
    \begin{tikzpicture}
    	    \node[inner sep=0] (image) {\includegraphics[width=.95\linewidth]{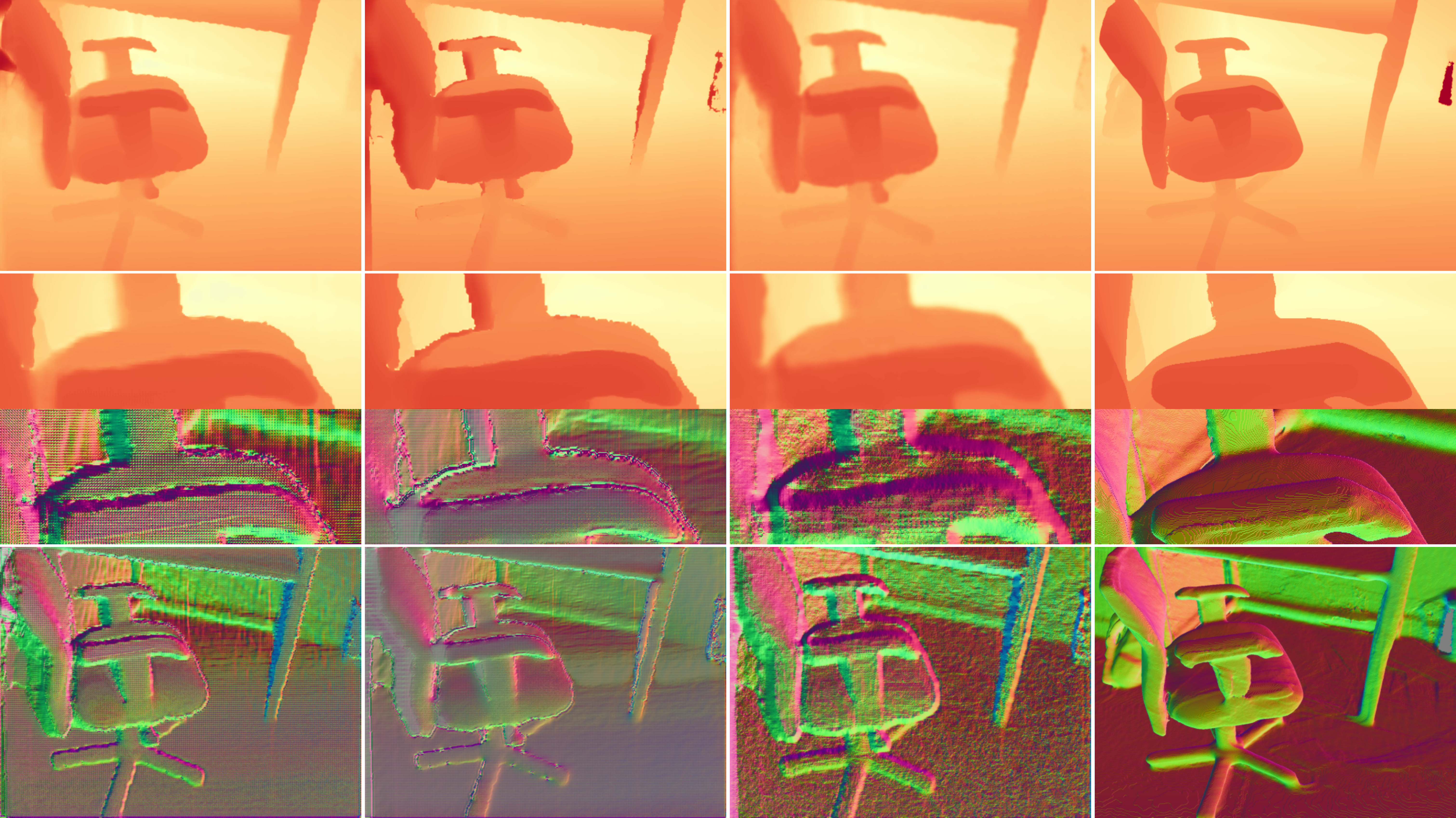}};
            \begin{scope}[shift=(image.north west), x={(image.north east)},y={(image.south west)}]
            \node at (.14, -.04) {\footnotesize {Gu+SRFBN}};
            \node at (.39, -.04) {\footnotesize {bicubic+Gu}};
            \node at (.63, -.04) {\footnotesize {UDSR}};
            \node at (.87, -.04) {\footnotesize {Reference ($\tgt$)}};
            \end{scope}
        \end{tikzpicture}
        \caption{Depth SR}
    	\label{fig:sr_interior}
	\end{subfigure}
    \caption{Enhancement and SR using the \textit{ScanNet-InteriorNet} benchmark.
    Our approach presents visually appealing results, particularly noticeable when observing normals. You can find additional results supplementary materials in ~\cref{fig:sup_enh_scannet_interiornet} and ~\cref{fig:sup_sr_scannet_interiornet}}
	\label{fig:enh}
\end{figure*}

\vspace{0.5em}
\noindent \textit{Results.}
The evaluation results presented in~\cref{tab:enh_scannet_interiornet} demonstrate significant quantitative performance gains across quality measures we compute, particularly for \mse[v].
Visual results~\cref{fig:enh_interior} demonstrate that our method recovers normals better compared to Gu~\etal, resulting in more visually appealing surface geometry.

\subsection{Ablation Studies}
\label{experiments:ablation}

\begin{figure*}
        \begin{tikzpicture}
            \node[inner sep=0] (image) {\includegraphics[width=.95\linewidth]{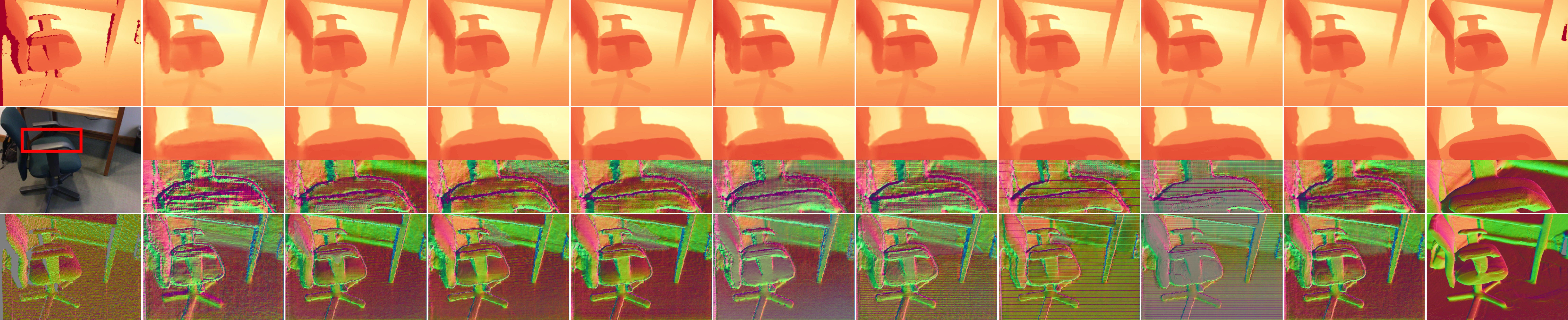}};
            \begin{scope}[shift=(image.north west), x={(image.north east)},y={(image.south west)}]
                \node[rotate=90] at (-.02, .14) {\footnotesize {Depth}};
                \node[rotate=90] at (-.025, .48) {\footnotesize {RGB}};
                \node[rotate=90] at (-.01, .48) {\footnotesize {Close-up}};
                \node[rotate=90] at (-.025, .82) {\footnotesize {Surface}};
                \node[rotate=90] at (-.01, .82) {\footnotesize {Normals}};
                \node at (.05, -.04) {\footnotesize {Input ($\src$)}};
                \node at (.15, -.04) {\footnotesize {$\text{UDSR}^{\bullet}$}};
                \node at (.23, -.04) {\footnotesize {$\text{UDSR}^\ddagger$}};
                \node at (.32, -.04) {\footnotesize {$\text{UDSR}^\ast$}};
                \node at (.41, -.04) {\footnotesize {$\text{UDSR}^\star$}};
                \node at (.50, -.04) {\footnotesize {$\text{UDSR}^\dagger$}};
                \node at (.59, -.04) {\footnotesize {$\text{UDSR}^\S$}};
                \node at (.68, -.04) {\footnotesize {$\text{UDSR}^\diamond$}};
                \node at (.77, -.04) {\footnotesize {$\text{UDSR}^\circ$}};
                \node at (.86, -.04) {\footnotesize {UDSR}};
                \node at (.955, -.04) {\footnotesize {Reference ($\tgt$)}};
            \end{scope}
    	\end{tikzpicture}
    	\label{fig:abl_sr_int}
    \caption{Ablation study of our depth SR method using data in \textit{ScanNet-InteriorNet} benchmark (see~\cref{tab:ablation_sr_interiornet} for notation). You can find additional results supplementary materials in ~\cref{fig:enh_ablation_int} and ~\cref{fig:sup_sr_ablation_int}}
\end{figure*}

We study the effects of different components of our method by training its several versions on \textit{ScanNet-InteriorNet} benchmark.
Results of this ablative study are presented quantitatively in~\cref{tab:ablation_sr_interiornet} and visually in~\cref{fig:abl_sr_int}.

Removing the RGB guidance network~$\fimg$ (UDSR$^\text{\textdagger}$) and training without augmenting input data using random gaps (UDSR$^{\diamond}$) impairs inpainting performance both qualitatively and quantitatively as assessed by \rmse[h] and \mae[h].
These findings are expected, as RGB guidance enables predicting depth in areas where direct range measurements are unavailable, while adding random holes is meant to provide robustness of our approach.

Compared to generating pseudo-examples using an unmodified CycleGAN (UDSR$^\text{\textbullet}$), our Translation component significantly improves performance for the full method \wrt~all measures. 
Training without pseudo-examples at all (UDSR$^{\circ}$) produces low-quality surfaces, both qualitatively and as measured by \mse[v].


\begingroup

\renewcommand{\arraystretch}{1.15}
\setlength\tabcolsep{\widthof{0}*\real{.75}}

\def\u{\underline}
\def\b{\textbf}

\begin{table}
\centerline{
\begin{threeparttable}
\begin{tabular}{l r r r r r}
\toprule
Method                  & 
    \rmse[h]            & 
    \rmse[d]            & 
    \mae[h]             & 
    \mae[d]             & 
    \mse[v]             \\ 
\midrule
UDSR\tnote{\textbullet}     & 238.4        & 203.8      & 183.7     & 188.2     & 39.3      \\
UDSR\tnote{$\ddagger$}      & 213.3        &  66.1      & 135.2     &  31.0     & 24.4      \\
UDSR\tnote{$\ast$}          & 208.6        &  69.3      & 135.0     &  36.3     & \u{23.6}  \\
UDSR\tnote{$\star$}         & \u{192.0}    &  63.1      & \u{119.4} &  25.4     & \b{23.0}  \\ 
\midrule
UDSR\tnote{\textdagger}     & 376.3        &  66.5      & 271.5     &  29.7     & 41.2      \\
UDSR\tnote{\textsection}    & 228.4        &  71.6      & 151.1     &  36.6     & 35.1      \\
UDSR\tnote{$\diamond$}      & 342.1        &  66.2      & 279.9     &  28.8     & 28.8      \\
UDSR\tnote{$\circ$}         & \b{177.7}    & \u{62.8}   & \b{104.2} & \u{24.6}  & 43.2      \\
UDSR                        & 197.9        & \b{61.0}   & 123.5     & \b{20.6}  & 26.1      \\
\bottomrule
\end{tabular}
\begin{tablenotes}
\scriptsize
    \item Modifications of the unpaired translation step:
    \item[\textbullet] replaced with unmodified CycleGAN,
    \item[$\ddagger$] trained without normals-based discriminators \(\discNmapTgt, \discNmapSrc\),
    \item[$\ast$] trained without \mse[v] in \(\Loss_{\text{cycle}}\),
    \item[$\star$] trained without depth range loss \(\Loss_{\text{range}}\).
    \item Modifications of the enhancement algorithm:
    \item[\textdagger] without RGB guidance network \(\fimg\),
    \item[\textsection] trained without normal-based loss term \(\Loss_{\text{surf}}\),
    \item[$\diamond$] trained without additional holes in the input,
    \item[$\circ$] trained without pseudo-examples.
\end{tablenotes}
\end{threeparttable}
}
\caption{Ablation study of our depth SR algorithm using the data in \textit{ScanNet-InteriorNet} benchmark.}
\label{tab:ablation_sr_interiornet}
\end{table}

\endgroup

\section{Conclusions}
\label{sec:conclusion}

In this work, we have demonstrated how data-driven depth super-resolution and enhancement can be performed without using paired datasets, which, in turn, simplifies training models for this task on real-world depth data.
We have constructed a learning-based pipeline for training a depth enhancement model using collections of data that do not depict the same real 3D environments and feature depth acquired at significantly different resolution and quality; we introduced a number of improvements into the pipeline and its stages, aiming to enable more robust, accurate, and complete depth reconstructions. 

We further proposed a benchmark for depth super-resolution based on real-world RGB-D scans available in existing collections~\cite{ScanNet:Dai2017} and higher quality data obained by rendering reconstructions obtained from multiple such scans. 

One important direction for future work is development of paired training and evaluation datasets with high-fidelity reference depth measurements, for more accurate validation and training~\cite{voynov2022multi,he2021towards}.  
Although our method outperformed existing unpaired methods, we believe that performance can be further improved with fine-tuning the models. 
The improvement of the model robustness and scalability is another possible future direction.

\section*{Acknowledgements}


This work was supported by Ministry of Science and Higher Education grant No. 075-10-2021-068.

\clearpage\newpage\FloatBarrier

\FloatBarrier


    \FloatBarrier
    {\clearpage\newpage\small
    \bibliographystyle{elsarticle-num} 
    \bibliography{references}}

\section{Details of Unpaired Translation Algorithm}
\label{supp:sec:translation}

For training the translation network, we use an adversarial loss formulation proposed by LSGAN~\cite{LSGAN:Mao2017}.
Specifically, for the depth- and normals-based discriminator networks $\discDmapSrc$ and $\discNmapSrc$ and the translation network $\GenTgtSrc$ we optimize
\begin{align}
\label{eq:translation-adv-losses}
\Loss_{\text{adv}} (\discDmapSrc)
=& \frac 1 2
    \mathbb{E}_{(\image, \dmap^{\src}) \sim \mathrm{P}_{\src}} 
    \left[(\discDmapSrc(\dmap^{\src}) -1)^2\right]
+ \frac 1 2 
    \mathbb{E}_{(\image, \dmap^{\tgtdown}) \sim \mathrm{P}_{\tgtdown}} 
    \left[(\discDmapSrc(\GenTgtSrc(\image, \dmap^{\tgtdown})))^2\right]  \nonumber \\ 
\Loss_{\text{adv}} (\discNmapSrc)
=& \frac 1 2
    \mathbb{E}_{(\image, \dmap^{\src}) \sim \mathrm{P}_{\src}} 
    \left[(\discNmapSrc(\tau(\dmap^{\src})) -1)^2\right]
+ \frac 1 2 
    \mathbb{E}_{(\image, \dmap^{\tgtdown}) \sim \mathrm{P}_{\tgtdown}} 
    \left[(\discNmapSrc(\tau(\GenTgtSrc(\image, \dmap^{\tgtdown}))))^2\right] \\
\Loss_{\text{adv}} (\GenTgtSrc)
=& \frac 1 2
    \mathbb{E}_{(\image, \dmap^{\tgtdown}) \sim \mathrm{P}_{\tgtdown}} 
    \left[(\discDmapSrc(\GenTgtSrc(\image, \dmap^{\tgtdown})) -1)^2\right] \nonumber \\ 
& + \frac 1 2
    \mathbb{E}_{(\image, \dmap^{\tgtdown}) \sim \mathrm{P}_{\tgtdown}} 
    \left[(\discNmapSrc(\tau(\GenTgtSrc(\image, \dmap^{\tgtdown}))) -1)^2\right].  \nonumber
\end{align}
In \eqref{eq:translation-adv-losses}, $\tau(\cdot)$ represents a differentiable operator used to compute the finite-difference estimates of per-pixel normals from the depth images, and the symbols $\mathrm{P}_{\src}$ and $\mathrm{P}_{\tgtdown}$ denote the distributions of the source $\src$ and the target $\tgtdown$ sets of RGB-D images, respectively.

\section{Details of Evaluation Methodology}
\label{supp:sec:benchmark}

\vspace{0.5em}

To prepare the data for our benchmark we used the following process.
To obtain the high-quality target depth maps for the \textit{ScanNet-RenderScanNet} and the \textit{Testing dataset,} we used an approach similar to the one proposed in~\cite{RecScanNet:Jeon2018}.
Specifically, we reconstructed 3D models of ScanNet scenes using BundleFusion~\cite{BundleFusion:Dai2017} and rendered the models using RGB camera parameters from the dataset, getting the renders paired with the respective RGB-D images.
In comparison to the noisy sensor depth maps with significant regions of missing values, the rendered depth maps are defined almost everywhere and have a noise level reduced by depth fusion.
Due to occasional errors in camera trajectory from the dataset and fusion artefacts, some pairs had a misalignment between the sensor depth map and the render.
To find such pairs and only keep the data with the best alignment, we computed values of the structural similarity index SSIM~\cite{SSIM:Zhou2004} between the sensor depth and the render.

Based on these values, we manually selected the best pairs of full-sized RGB-D images and respective renders without misalignment for the \textit{Testing dataset.}
For a better evaluation of the ability of methods to fill in missing values,
we additionally filtered out some pairs so that the distribution of areas of missing values in the sensor depth maps in the \textit{Testing dataset} would match the raw data distribution in ScanNet.

Then, we split the ScanNet scenes that did not get into the \textit{Testing dataset} randomly between Train A, Train B, and Val sub-parts of ScanNet-RenderScanNet.
We included full-sized sensor RGB-D images in Train A.
To get a larger number of samples in Train B and Val and not reject all the data from images with only a small region with artefacts, we further split the data into pairs of crops from sensor RGB-D images with \(320\times 320\) depth
and crops from renders of size \(640\times 640\).
After that, we only kept the crop pairs with SSIM between the sensor depth and the render higher than 0.8.


As the source of high-quality data for \textit{ScanNet-InteriorNet} we used RGB-D images from InteriorNet.
Depth images in InteriorNet have zero level of noise and no missing values, but their resolution is twice lower than the resolution of RGB images.
To obtain the depth maps of the same resolution as RGB images, we used a 2x super-resolution method SRFBN~\cite{SRFBN:Li_2019} that we trained on the high-quality renders of ScanNet.



\section{Training Details}
\label{supp:sec:training}

\begin{table}[]
\centering
\resizebox{\textwidth}{!}{%
\begin{tabular}{@{}lrllllllllllll@{}}
\toprule
 & \multicolumn{1}{l}{} &  &  & \multicolumn{6}{c}{Opt. Params} & \multicolumn{4}{c}{Inputs} \\ \midrule
Scenario & \multicolumn{1}{l}{Network} & Loss Function & Weights/Masks & BS & LR & NI & $w_{\text{decay}}$ & $\beta_1$ & $\beta_2$ & Set & $N$ & Res. & Aug. \\ \midrule
\begin{tabular}[c]{@{}l@{}}\textit{ScanNet-}\\ \textit{RenderScanNet}\end{tabular} & \multicolumn{1}{l}{\begin{tabular}[c]{@{}l@{}}Translation \\ Network (stage 1)\end{tabular}} & \begin{tabular}[c]{@{}l@{}}$\Loss_{\text{adv}} + \lambda_{\text{cycle}} \Loss_{\text{cycle}}$\\ $ + \lambda^{\src}_{\text{range}} \Reg^{\src}_{\text{range}}$\\ $ + \lambda^{\tgtdown}_{\text{range}} \Reg^{\tgtdown}_{\text{range}}$\\ $ + \lambda_{\text{idt}} \Reg_{\text{idt}}$ \eqref{eq:full_translation_objective}\end{tabular} & \begin{tabular}[c]{@{}l@{}}$\lambda_{\text{cycle}} = 5$\\ $\lambda^{\src}_{\text{range}} = 2$\\ $\lambda^{\tgtdown}_{\text{range}} = 1$\\ $\lambda_{\text{idt}} = 1$\end{tabular} & 6 & $2\cdot 10^{-4}$ & 120K & $10^{-4}$ & 0.5 & 0.999 & $\src, \tgtdown$ & 37986 & $320 \times 320$ & rnd. crop $256 \times 256$ \\
 & \textit{(stage 2)} &  &  &  & lin. reduce to 0 & 175K &  &  &  &  &  &  &  \\ \cmidrule(l){2-14} 
 & \multicolumn{1}{l}{\begin{tabular}[c]{@{}l@{}}RGB Guidance\\ Network (stage 1)\end{tabular}} & Masked $L_1$ loss & \begin{tabular}[c]{@{}l@{}}Exclude missing\\ pixels in input depth\end{tabular} & 12+12 & $2\cdot 10^{-3}$ & 150K & 0 & 0.9 & 0.999 & $\src, \tgtdown$ & 37986 & $640 \times 480$ & \begin{tabular}[c]{@{}l@{}}rnd. crop $512 \times 384$\\ rnd. rotations\\ horizontal flips\end{tabular} \\
 & \textit{(stage 2)} & \textit{} &  &  & lin. reduce to 0 & 150K &  & \textit{} &  &  &  &  &  \\ \cmidrule(l){2-14} 
 & \multicolumn{1}{l}{\begin{tabular}[c]{@{}l@{}}Enhancement \\ Network (stage 1)\end{tabular}} & \begin{tabular}[c]{@{}l@{}}$\bm{\lambda}_{\text{depth}}^{\pseudosrc} (\bm{L}_{\text{depth}}^{\pseudosrc})^{\intercal}$\\ $ + \bm{\lambda}_{\text{surf}}^{\pseudosrc} (\bm{L}_{\text{surf}}^{\pseudosrc})^{\intercal}$\\ $ + \lambda^{\pseudosrc}_{\smooth} \RegSmooth^{\pseudosrc}$\\ $ + \bm{\lambda}_{\text{depth}}^{\src} (\bm{L}_{\text{depth}}^{\src})^{\intercal}$\\ $ + \lambda_{\edge}^{\src} \RegEdge^{\src}$\\ $ + \lambda_{\smooth}^{\src} \RegSmooth^{\src}$ \eqref{eq:enh_full_objective}\end{tabular}  & \begin{tabular}[c]{@{}l@{}}$\bm{\lambda}_{\text{depth}}^{\pseudosrc} = \begin{bmatrix}15 & 10\end{bmatrix}$\\ $\bm{\lambda}_{\text{surf}}^{\pseudosrc} = \begin{bmatrix}3 & 3\end{bmatrix}$\\ $\lambda^{\pseudosrc}_{\smooth} = 2 \cdot 10^{-7}$\\ $\bm{\lambda}_{\text{depth}}^{\src} = \begin{bmatrix}40 & 20\end{bmatrix}$\\ $\lambda^{\pseudosrc}_{\smooth} = 2 \cdot 10^{-7}$\\ $\lambda^{\pseudosrc}_{\edge} = 1$\\ $\bm{w}_{\text{depth}, 1}^{\pseudosrc} = \begin{bmatrix}1 & 30\end{bmatrix}$\\ $\bm{w}_{\text{depth}, 2}^{\pseudosrc} = \begin{bmatrix}0 & 20\end{bmatrix}$\\ $\bm{w}_{\text{depth}, 1}^{\src} = \begin{bmatrix}1 & 40\end{bmatrix}$\\ $\bm{w}_{\text{depth}, 2}^{\src} = \begin{bmatrix}0 & 20\end{bmatrix}$\end{tabular} & 12 & $10^{-3}$ & 20K & 0 & 0.9 & 0.999 & $\src, \tgtdown$ &  & $640 \times 480$ & \begin{tabular}[c]{@{}l@{}}rnd. crop $512 \times 384$\\ rnd. rotations\\ horizontal flips\end{tabular} \\
 & \textit{(stage 2)} &  &  &  & lin. reduce to $4 \cdot 10^{-4}$ & 60K &  &  &  &  &  &  &  \\
 & \textit{(stage 3)} &  & \begin{tabular}[c]{@{}l@{}}$\bm{w}_{\text{depth}, 1}^{\pseudosrc} = \begin{bmatrix}1 & 120\end{bmatrix}$\\ $\bm{w}_{\text{depth}, 2}^{\pseudosrc} = \begin{bmatrix}0 & 60\end{bmatrix}$\\ $\bm{w}_{\text{depth}, 1}^{\src} = \begin{bmatrix}2 & 40\end{bmatrix}$\\ $\bm{w}_{\text{depth}, 2}^{\src} = \begin{bmatrix}0 & 50\end{bmatrix}$\end{tabular} & 6 & $2 \cdot 10^{-4}$ & 10K &  &  &  &  &  &  & no rnd. crop \\
 & \textit{(stage 4)} &  &  &  & lin. reduce to 0 & 20K &  &  &  &  &  &  &  \\
 & \textit{(adapt for SR 1)} &  & $\bm{\lambda}_{\text{surf}}^{\pseudosrc} = \begin{bmatrix}4 & 4\end{bmatrix}$ &  & $2 \cdot 10^{-4}$ & 5K &  &  &  & $\srcup, \tgt$ &  &  & rnd. crop $384 \times 384$ \\
 & \textit{(adapt for SR 2)} &  &  &  & lin. reduce to 0 & 25K &  &  &  &  &  &  &  \\ \midrule
\begin{tabular}[c]{@{}l@{}}\textit{ScanNet-}\\ \textit{InteriorNet}\end{tabular} & \multicolumn{1}{l}{\begin{tabular}[c]{@{}l@{}}Translation \\ Network (stage 1)\end{tabular}} & \begin{tabular}[c]{@{}l@{}}$\Loss_{\text{adv}} + \lambda_{\text{cycle}} \Loss_{\text{cycle}}$\\ $ + \lambda^{\src}_{\text{range}} \Reg^{\src}_{\text{range}}$\\ $ + \lambda^{\tgtdown}_{\text{range}} \Reg^{\tgtdown}_{\text{range}}$\\ $ + \lambda_{\text{idt}} \Reg_{\text{idt}}$ \eqref{eq:full_translation_objective}\end{tabular} & \begin{tabular}[c]{@{}l@{}}$\lambda_{\text{cycle}} = 2$\\ $\lambda^{\src}_{\text{range}} = 2$\\ $\lambda^{\tgtdown}_{\text{range}} = 2$\\ $\lambda_{\text{idt}} = 1$\end{tabular} & 6 & $2 \cdot 10^{-4}$ & 114K & $10^{-4}$ & 0.5 & 0.999 & $\src, \tgtdown$ & 37986 & $320 \times 320$ & rnd. crop $256 \times 256$ \\
 & \textit{(stage 2)} &  &  &  & lin. reduce to 0 & 204K &  &  &  &  &  &  &  \\ \cmidrule(l){2-14} 
 & \multicolumn{1}{l}{\begin{tabular}[c]{@{}l@{}}Enhancement \\ Network (stage 1)\end{tabular}} & \begin{tabular}[c]{@{}l@{}}$\bm{\lambda}_{\text{depth}}^{\pseudosrc} (\bm{L}_{\text{depth}}^{\pseudosrc})^{\intercal}$\\ $ + \bm{\lambda}_{\text{surf}}^{\pseudosrc} (\bm{L}_{\text{surf}}^{\pseudosrc})^{\intercal}$\\ $ + \lambda^{\pseudosrc}_{\smooth} \RegSmooth^{\pseudosrc}$\\ $ + \bm{\lambda}_{\text{depth}}^{\src} (\bm{L}_{\text{depth}}^{\src})^{\intercal}$\\ $ + \lambda_{\edge}^{\src} \RegEdge^{\src}$\\ $ + \lambda_{\smooth}^{\src} \RegSmooth^{\src}$ \eqref{eq:enh_full_objective}\end{tabular}  & \begin{tabular}[c]{@{}l@{}}$\bm{\lambda}_{\text{depth}}^{\pseudosrc} = \begin{bmatrix}10 & 10\end{bmatrix}$\\ $\bm{\lambda}_{\text{surf}}^{\pseudosrc} = \begin{bmatrix}3 & 3\end{bmatrix}$\\ $\lambda^{\pseudosrc}_{\smooth} = 2 \cdot 10^{-7}$\\ $\bm{\lambda}_{\text{depth}}^{\src} = \begin{bmatrix}40 & 20\end{bmatrix}$\\ $\lambda^{\pseudosrc}_{\smooth} = 2 \cdot 10^{-7}$\\ $\lambda^{\pseudosrc}_{\edge} = 1$\\ $\bm{w}_{\text{depth}, 1}^{\pseudosrc} = \begin{bmatrix}1 & 40\end{bmatrix}$\\ $\bm{w}_{\text{depth}, 2}^{\pseudosrc} = \begin{bmatrix}0 & 25\end{bmatrix}$\\ $\bm{w}_{\text{depth}, 1}^{\src} = \begin{bmatrix}1 & 40\end{bmatrix}$\\ $\bm{w}_{\text{depth}, 2}^{\src} = \begin{bmatrix}0 & 25\end{bmatrix}$\end{tabular} & 12 & $10^{-3}$ & 20K & 0 & 0.9 & 0.999 & $\src, \tgtdown$ &  &  & \begin{tabular}[c]{@{}l@{}}rnd. crop $512 \times 384$\\ rnd. rotations\\ horizontal flips\end{tabular} \\
 & \textit{(stage 2)} &  &  &  & lin. reduce to 0 & 80K &  &  &  &  &  &  &  \\
 & \textit{(stage 3)} &  &  & 6 & $2 \cdot 10^{-4}$ & 10K &  &  &  &  &  &  & no rnd. crop \\
 & \textit{(stage 4)} &  &  &  & lin. reduce to 0 & 20K &  &  &  &  &  &  &  \\
 & \textit{(adapt for SR 1)} &  & \begin{tabular}[c]{@{}l@{}}$\bm{\lambda}_{\text{surf}}^{\pseudosrc} = \begin{bmatrix}4 & 4\end{bmatrix}$\\ $\lambda^{\pseudosrc}_{\smooth} = 10^{-7}$\\ $\lambda^{\pseudosrc}_{\smooth} = 10^{-7}$\\ $\bm{w}_{\text{depth}, 1}^{\pseudosrc} = \begin{bmatrix}1 & 60\end{bmatrix}$\\ $\bm{w}_{\text{depth}, 2}^{\pseudosrc} = \begin{bmatrix}0 & 70\end{bmatrix}$\end{tabular} &  & $2 \cdot 10^{-4}$ & 5K &  &  &  & $\srcup, \tgt$ &  &  & \begin{tabular}[c]{@{}l@{}}rnd. crop $384 \times 384$\\ rnd. rotations\\ horizontal flips\end{tabular} \\
 & \textit{(adapt for SR 2)} &  &  &  & lin. reduce to 0 & 30K &  &  &  &  &  &  &  \\
 & \textit{(adapt for SR 3)} &  &  &  & $10^{-4}$ & 5K &  &  &  &  &  &  & no augs \\
 & \textit{(adapt for SR 4)} &  &  &  & lin. reduce to 0 & 10K &  &  &  &  &  &  & \\ \bottomrule
\end{tabular}%
}
\caption{BS denotes batch size, LR denotes learning rate, NI denotes number of iterations used during the optimization.}
\label{tab:training-details-all}
\end{table}

We train our models in several stages for each of the training scenarios (\textit{ScanNet-RenderScanNet}) and \textit{ScanNet-InteriorNet}), where we specify a training schedule (learning rate, optimization hyperparameters) as well as our optimization objective (choice of weights for the loss) on a per-stage basis.
We give a summary of our training details, including the choice of data, weighting, and optimization settings in Table~\ref{tab:training-details-all}. 
We review details specific to training each network in respective paragraphs.

\vspace{0.5em}
\noindent \textit{Training the Translation Networks.}
To train our translation algorithm, we initialize parameters for all networks using Xavier initialization~\cite{Xavier:Glorot2010} and use Adam optimizer~\cite{Adam:Kingma2015} to find the optimal weights for these networks.
During training, we update generators three times per each discriminator update.





\vspace{0.5em}
\noindent \emph{Training the RGB Guidance Network.}
We train $\fimg$ using Adam optimizer~\cite{Adam:Kingma2015} (we additionally set $\epsilon=10^{-8}$), forming each batch of 24 RGB-D images using 12~low-quality and 12~high-quality RGB-D images.

\begin{figure*}
        \begin{tikzpicture}
            \node[inner sep=0] (image) {\includegraphics[width=.95\linewidth]{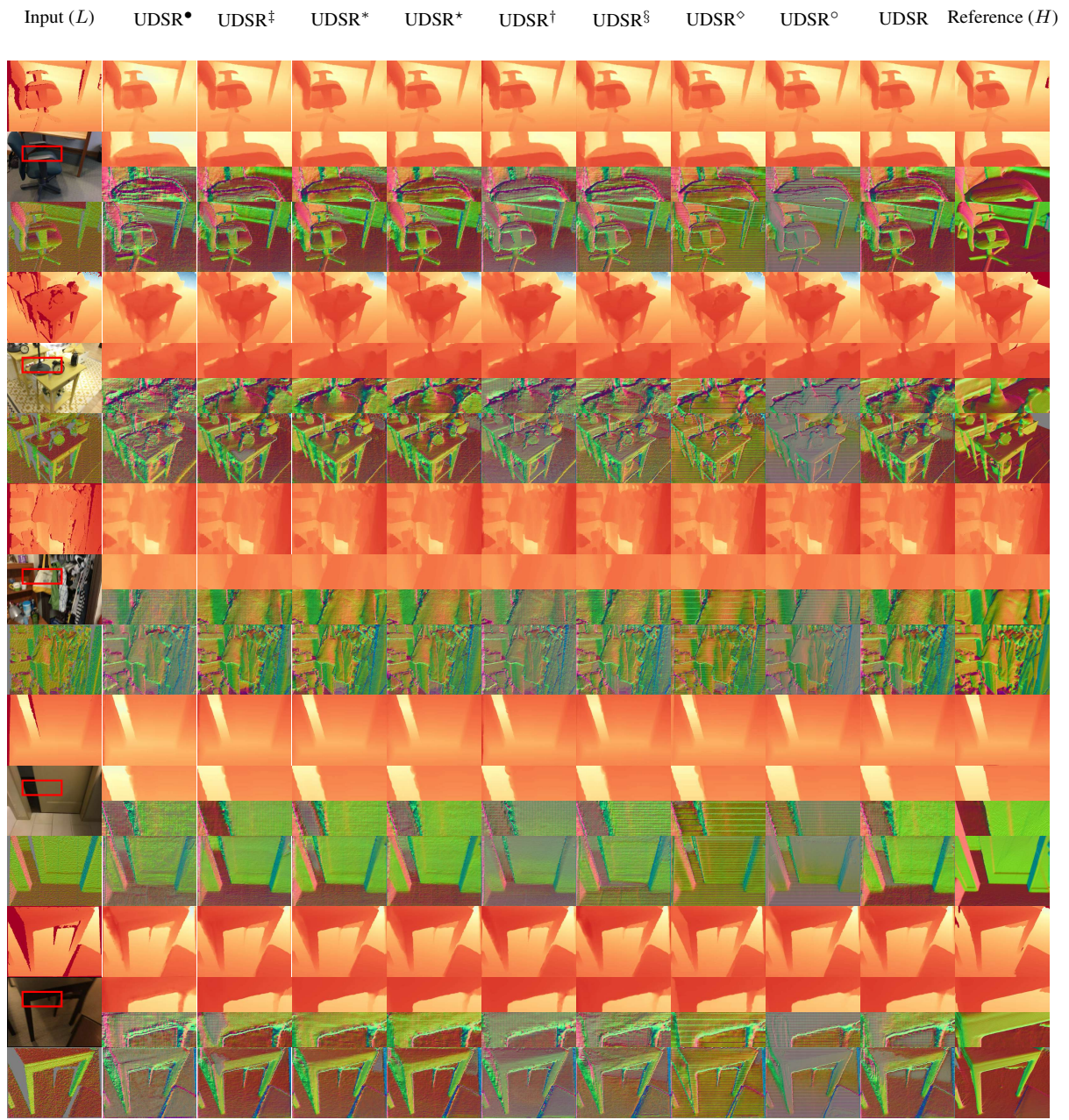}};
            \begin{scope}[shift=(image.north west), x={(image.north east)},y={(image.south west)}]
                \node at (.05, -.04) {\footnotesize {Input ($\src$)}};
                \node at (.15, -.04) {\footnotesize {$\text{UDSR}^{\bullet}$}};
                \node at (.23, -.04) {\footnotesize {$\text{UDSR}^\ddagger$}};
                \node at (.32, -.04) {\footnotesize {$\text{UDSR}^\ast$}};
                \node at (.41, -.04) {\footnotesize {$\text{UDSR}^\star$}};
                \node at (.50, -.04) {\footnotesize {$\text{UDSR}^\dagger$}};
                \node at (.59, -.04) {\footnotesize {$\text{UDSR}^\S$}};
                \node at (.68, -.04) {\footnotesize {$\text{UDSR}^\diamond$}};
                \node at (.77, -.04) {\footnotesize {$\text{UDSR}^\circ$}};
                \node at (.86, -.04) {\footnotesize {UDSR}};
                \node at (.955, -.04) {\footnotesize {Reference ($\tgt$)}};
            \end{scope}
    	\end{tikzpicture}
    \caption{Ablation of SR on \textbf{ScanNet-InteriorNet}. Same notation as in Table~\ref{tab:ablation_sr_interiornet}.}
	\label{fig:sup_sr_ablation_int}
\end{figure*}

\begin{figure*}
        \begin{tikzpicture}
            \node[inner sep=0] (image) {\includegraphics[width=.95\linewidth]{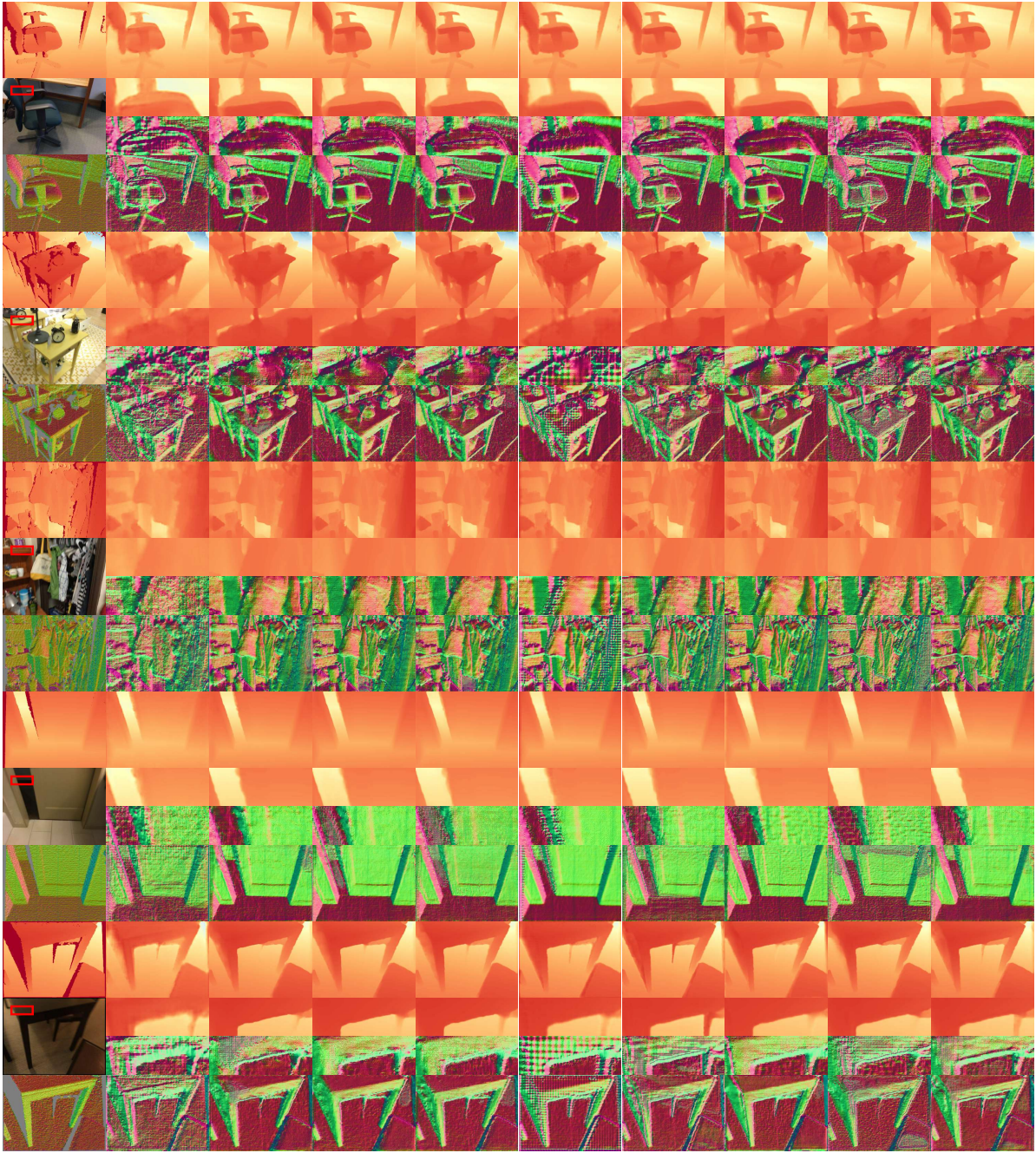}};
            \begin{scope}[shift=(image.north west), x={(image.north east)},y={(image.south west)}]
                \node at (.05, -.04) {\footnotesize {Input ($\src$)}};
                \node at (.15, -.04) {\footnotesize {$\text{UDSR}^{\bullet}$}};
                \node at (.23, -.04) {\footnotesize {$\text{UDSR}^\ddagger$}};
                \node at (.32, -.04) {\footnotesize {$\text{UDSR}^\ast$}};
                \node at (.41, -.04) {\footnotesize {$\text{UDSR}^\star$}};
                \node at (.50, -.04) {\footnotesize {$\text{UDSR}^\dagger$}};
                \node at (.59, -.04) {\footnotesize {$\text{UDSR}^\S$}};
                \node at (.68, -.04) {\footnotesize {$\text{UDSR}^\diamond$}};
                \node at (.77, -.04) {\footnotesize {$\text{UDSR}^\circ$}};
                \node at (.86, -.04) {\footnotesize {UDSR}};
                \node at (.955, -.04) {\footnotesize {Reference ($\tgt$)}};
            \end{scope}
    	\end{tikzpicture}
    \caption{Ablation of Enhancement on \textbf{ScanNet-InteriorNet}. Same notation as in Table~\ref{tab:ablation_sr_interiornet}.}
	\label{fig:enh_ablation_int}
\end{figure*}

\begin{figure*}
    \begin{center}
        \begin{tikzpicture}
            \node[inner sep=0] (image) {\includegraphics[width=0.95\textwidth,height=0.95\textheight,keepaspectratio]{img/s_enhancement_render.pdf}};
            \begin{scope}[shift=(image.north west), x={(image.north east)},y={(image.south west)}]
            \node at (.06, -.01) {\footnotesize {Input ($\src$)}};
            \node at (.17, -.01) {\footnotesize {$\text{CycleGAN}^{\text{\textbullet}}$}};
            \node at (.28, -.01) {\footnotesize {$\text{CycleGAN}^{\text{\textdagger}}$}};
            \node at (.39, -.01) {\footnotesize {$\text{U-GAT-IT}$}};
            \node at (.50, -.01) {\footnotesize {$\text{NiceGAN}$}};
            \node at (.61, -.01) {\footnotesize {Gu}};
            \node at (.72, -.01) {\footnotesize {LapDEN}};
            \node at (.83, -.01) {\footnotesize {Ours}};
            \node at (.94, -.01) {\footnotesize {Reference ($\tgt$)}};
            \end{scope}
        \end{tikzpicture}
    \end{center}
    \caption{Qualitative depth enhancement results for the \textbf{ScanNet-RenderScanNet} scenario.}
    \label{fig:sup_enh_scannet_rscannet}
\end{figure*}

\begin{figure*}
    \begin{center}
        \begin{tikzpicture}
            \node[inner sep=0] (image) {\includegraphics[width=0.95\textwidth,height=0.95\textheight,keepaspectratio]{img/s_enhancement_interior.pdf}};
            \begin{scope}[shift=(image.north west), x={(image.north east)},y={(image.south west)}]
            \node at (.06, -.01) {\footnotesize {Input ($\src$)}};
            \node at (.19, -.01) {\footnotesize {$\text{CycleGAN}^{\text{\textbullet}}$}};
            \node at (.31, -.01) {\footnotesize {$\text{CycleGAN}^{\text{\textdagger}}$}};
            \node at (.43, -.01) {\footnotesize {$\text{NiceGAN}$}};
            \node at (.56, -.01) {\footnotesize {$\text{U-GAT-IT}$}};
            \node at (.68, -.01) {\footnotesize {Gu}};
            \node at (.81, -.01) {\footnotesize {Ours}};
            \node at (.94, -.01) {\footnotesize {Reference ($\tgt$)}};
            \end{scope}
        \end{tikzpicture}
    \end{center}
    \caption{Qualitative depth enhancement results for \textbf{ScanNet-InteriorNet} scenario.}
    \label{fig:sup_enh_scannet_interiornet}
\end{figure*}
\begin{figure*}
    \begin{center}
        \begin{tikzpicture}
            \node[inner sep=0] (image) {\includegraphics[width=0.95\textwidth,height=0.95\textheight,keepaspectratio]{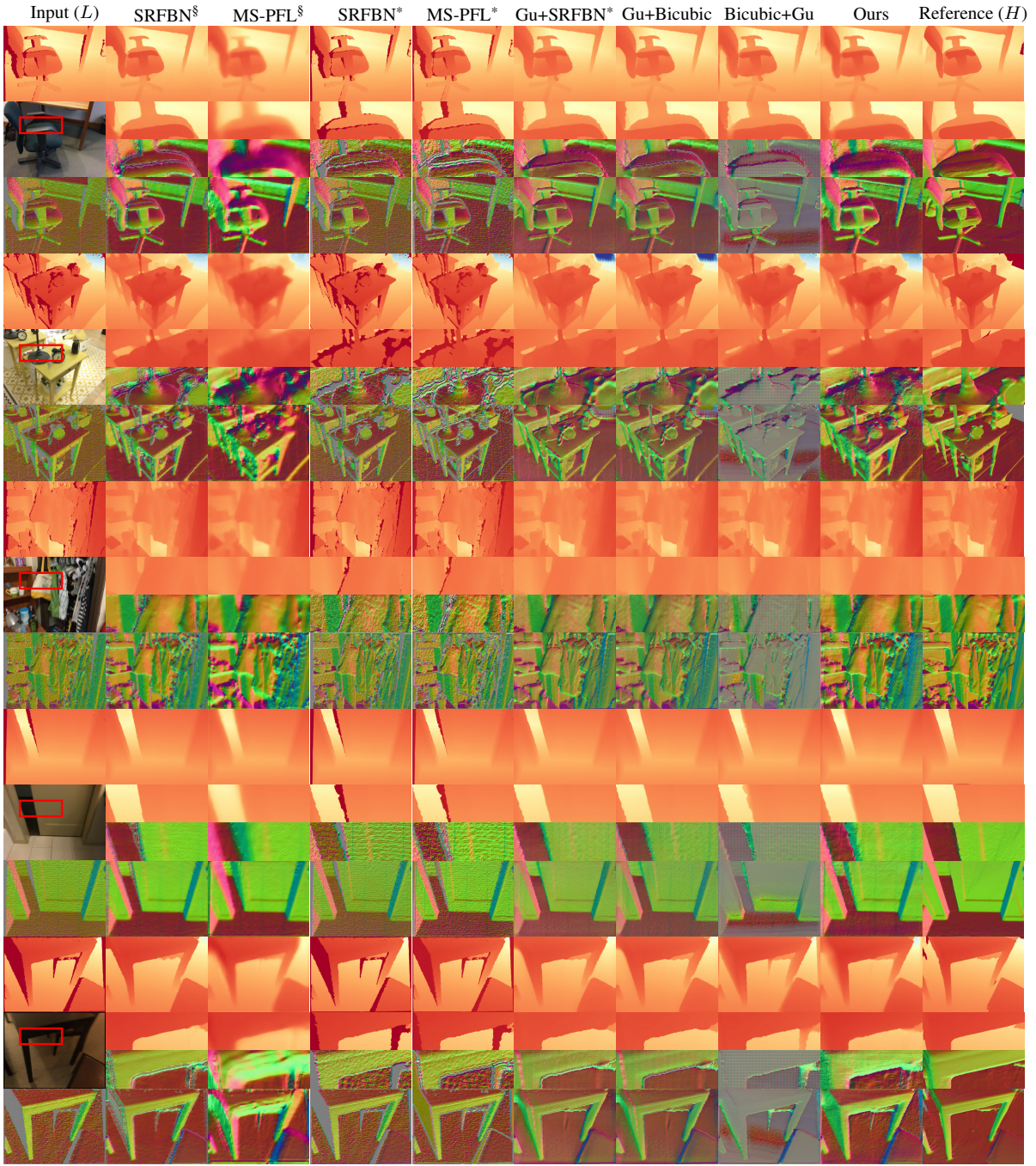}};
            \begin{scope}[shift=(image.north west), x={(image.north east)},y={(image.south west)}]
            \node at (.06, -.01) {\footnotesize {Input ($\src$)}};
            \node at (.16, -.01) {\footnotesize {$\text{SRFBN}^\text{\textsection}$}};
            \node at (.26, -.01) {\footnotesize {$\text{MS-PFL}^{\text{\textsection}}$}};
            \node at (.36, -.01) {\footnotesize {$\text{SRFBN}^\text{\textasteriskcentered}$}};
            \node at (.45, -.01) {\footnotesize {$\text{MS-PFL}^{\text{\textasteriskcentered}}$}};
            \node at (.55, -.01) {\footnotesize {Gu+$\text{SRFBN}^\text{\textasteriskcentered}$}};
            \node at (.65, -.01) {\footnotesize {Gu+Bicubic}};
            \node at (.75, -.01) {\footnotesize {Bicubic+Gu}};
            \node at (.85, -.01) {\footnotesize {Ours}};
            \node at (.95, -.01) {\footnotesize {Reference ($\tgt$)}};
            \end{scope}
        \end{tikzpicture}
    \end{center}
    \caption{Qualitative results for depth SR for \textbf{ScanNet-RenderScanNet} scenario.}
    \label{fig:sup_sr_scannet_rscannet}
\end{figure*}

\begin{figure*}
    \begin{center}
        \begin{tikzpicture}
            \node[inner sep=0] (image) {\includegraphics[width=0.95\textwidth,height=0.95\textheight,keepaspectratio]{img/s_sr_interior.pdf}};
            \begin{scope}[shift=(image.north west), x={(image.north east)},y={(image.south west)}]
            \node at (.08, -.01) {\footnotesize {Input ($\src$)}};
            \node at (.27, -.01) {\footnotesize {Gu+$\text{SRFBN}^\text{\textasteriskcentered}$}};
            \node at (.42, -.01) {\footnotesize {Gu+Bicubic}};
            \node at (.59, -.01) {\footnotesize {Bicubic+Gu}};
            \node at (.75, -.01) {\footnotesize {Ours}};
            \node at (.92, -.01) {\footnotesize {Reference ($\tgt$)}};
            \end{scope}
        \end{tikzpicture}
    \end{center}
    \caption{Qualitative depth SR results for \textbf{ScanNet-InteriorNet} scenario.}
    \label{fig:sup_sr_scannet_interiornet}
\end{figure*}
 
\end{document}